\documentclass[10pt,twocolumn,letterpaper]{article}

\usepackage[pagenumbers]{cvpr} 
\usepackage[accsupp]{axessibility}
\usepackage[dvipsnames]{xcolor}

\definecolor{cvprblue}{rgb}{0.21,0.49,0.74}
\definecolor{purple}{rgb}{0.33, 0.0, 0.49}

\usepackage[pagebackref,breaklinks,colorlinks,citecolor=cvprblue]{hyperref}
\usepackage{booktabs}
\usepackage{amsmath}
\usepackage{epsfig}
\usepackage{multirow}
\usepackage{ dsfont }
\usepackage{ cite }
\usepackage{float}
\usepackage{color, colortbl, xcolor}
\usepackage{soul}
\usepackage{float}
\usepackage{url}
\usepackage{afterpage}
\usepackage{rotate}
\usepackage{makecell}
\usepackage{multirow}
\usepackage{algorithm}
\usepackage{algorithmicx}
\usepackage{algpseudocode}
\usepackage{mathtools}
\usepackage{etoolbox}
\usepackage{lipsum}
\usepackage{bbm}
\usepackage{listings} 
\usepackage{tabularx}
\usepackage{amssymb}
\usepackage{amsbsy}
\usepackage{pifont} 
\usepackage{tikz}
\usetikzlibrary{shapes.geometric}
\usepackage[normalem]{ulem}

\usepackage{array}
\usepackage{transparent}

\definecolor{2nd}{gray}{0.8}

\newcolumntype{g}{>{\columncolor{Gray}}c}

\newcolumntype{^}{>{\currentrowstyle}}

\algdef{SE}[SUBALG]{Indent}{EndIndent}{}{\algorithmicend\ }%
\algtext*{Indent}
\algtext*{EndIndent}

\newcolumntype{Y}{>{\centering\arraybackslash}X}
\newcolumntype{P}[1]{>{\centering\arraybackslash}p{#1}}

\algnewcommand{\nComment}[1]{\Statex \Comment{#1}}

\definecolor{LightCyan}{rgb}{0.88,1,1}

\definecolor{mypurple}{RGB}{153, 0, 153}
\definecolor{mygray}{RGB}{128, 128, 128}
\definecolor{mygreen}{RGB}{0, 153, 0}

\definecolor{mycyan}{RGB}{64, 128, 128}
\definecolor{mypink}{RGB}{255, 182, 193}
\definecolor{myred}{RGB}{165,42,42}
\definecolor{myyellow}{RGB}{255, 191, 0}

\definecolor{tab_red}{rgb}{0.71, 0.11, 0.0}
\definecolor{tab_green}{rgb}{0.11, 0.71, 0.0}

\newcommand{\improve}[1]{\textcolor{tab_green}{\textbf{#1}}}
\newcommand{\drop}[1]{\textcolor{tab_red}{\textbf{#1}}}

\definecolor{car_yellow}{RGB}{255,255,153}
\definecolor{dog_purple}{RGB}{204,153,255}
\definecolor{bg_white}{RGB}{255,255,255}

\usetikzlibrary{shadows}
\usetikzlibrary{plotmarks}
\usetikzlibrary{shapes}

\newcommand{\thickhline}{%
	\noalign {\ifnum 0=`}\fi \hrule height 1pt
	\futurelet \reserved@a \@xhline
}
\makeatletter
\global\let\oriCT@@do@color\CT@@do@color

\def\paperID{2335} 
\def\confName{CVPR}
\def\confYear{2024}

\title{	Unraveling Instance Associations: A Closer Look for Audio-Visual Segmentation }

\author{
\parbox{0.7\linewidth}{\centering
Yuanhong Chen \textsuperscript{\rm 1} $\thanks{First two authors contributed equally to this work.}$ $\quad$
Yuyuan Liu \textsuperscript{\rm 1} \footnotemark[1] $\quad$
Hu Wang \textsuperscript{\rm 1} $\quad$
Fengbei Liu \textsuperscript{\rm 1} $\quad$
Chong Wang \textsuperscript{\rm 1} $\quad$
Helen Frazer\textsuperscript{\rm 2} $\quad$
Gustavo Carneiro\textsuperscript{\rm 3} $\newline$ 
\textsuperscript{\rm 1} Australian Institute for Machine Learning, University of Adelaide \\
\textsuperscript{\rm 2} St Vincent's Hospital Melbourne \\
\textsuperscript{\rm 3} Centre for Vision, Speech and Signal Processing, University of Surrey}
}

\begin{document}
\maketitle

\begin{abstract}
Audio-visual segmentation (AVS) is a challenging task that involves accurately segmenting sounding objects based on audio-visual cues. The effectiveness of audio-visual learning critically depends on achieving accurate cross-modal alignment between sound and visual objects.
Successful audio-visual learning requires two essential components: 1) a challenging dataset with high-quality pixel-level multi-class annotated images associated with audio files, and 2) a model that can establish strong links between audio information and its corresponding visual object.
However, these requirements are only partially addressed by current methods, with training sets containing biased audio-visual data, and models that generalise poorly beyond this biased training set. 
In this work, we propose a new cost-effective strategy to build challenging and relatively unbiased high-quality audio-visual segmentation benchmarks. 
We also propose a new informative sample mining method for audio-visual supervised contrastive learning to leverage discriminative contrastive samples to enforce cross-modal understanding.
We show empirical results that demonstrate the effectiveness of our benchmark. Furthermore, experiments conducted on existing AVS datasets and on our new benchmark show that our method achieves state-of-the-art (SOTA) segmentation accuracy\footnote{This work was supported by Australian Research Council through grant FT190100525.}. 
\end{abstract}

\vspace{-11pt}
\section{Introduction}
\label{sec:intro}

The human nervous system exhibits multi-modal perception~\cite{small2005odor}, combining input signals from different modalities to improve the detection and classification of multiple stimuli~\cite{small2005odor}.
Such functionality has been emulated by recent papers~\cite{arandjelovic2017look,arandjelovic2018objects,afouras2020self,chen2021localizing,mo2022localizing,hu2022mix,mo2022closer} that aim to associate visual objects with their corresponding audio sequences, in a task known as audio-visual correspondence (AVC)~\cite{arandjelovic2017look,arandjelovic2018objects}.

\begin{figure}
    \centering
    \includegraphics[width=1\linewidth]{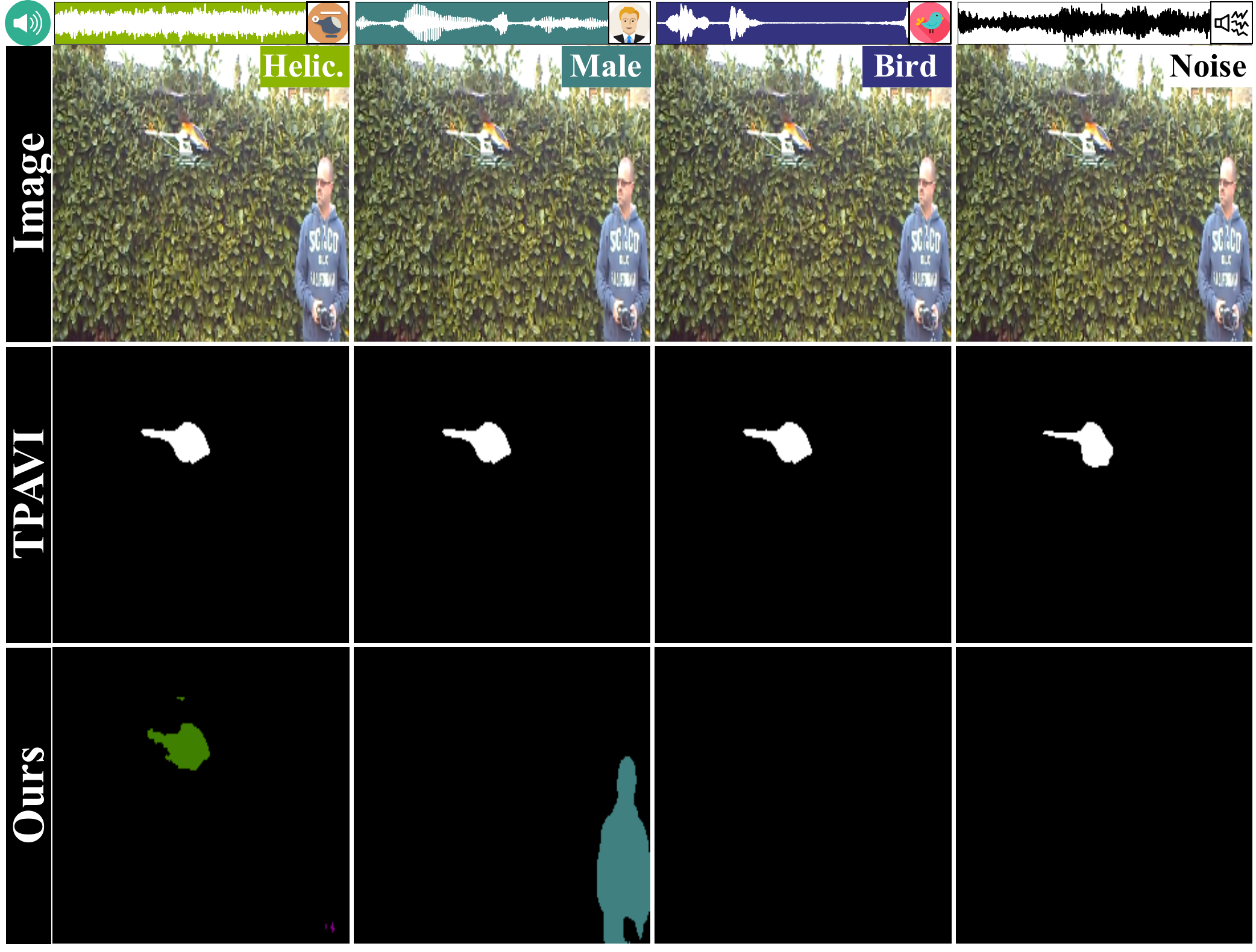}
    \vspace{-10pt}
    \caption{
    Current AVS datasets~\cite{zhou2022audio} tend to assume specific objects as consistent sound sources. Such a bias influences AVS methods, like TPAVI~\cite{zhou2022audio} (2nd row), to favour segmenting the presumed sound source, even when replacing the original audio with different sound types such as a person speaking (2nd column), bird chirping (3rd column), or background noise (4th row). Our paper proposes a new cost-effective strategy to build a relatively unbiased audio-visual segmentation benchmark and a supervised contrastive learning method that mines informative samples to better constrain the learning of audio-visual embeddings (last row). 
    }
    \vspace{-15pt}
    \label{fig:motivation}
\end{figure}

A particularly interesting AVC task is the audio-visual segmentation (AVS)~\cite{zhou2022audio, zhou2023audio} that aims to segment all pixels of the sounding visual objects using a fully supervised model. 
A major challenge in AVS is achieving cross-modal alignment between sound and visual objects~\cite{senocak2023sound}. 
Current datasets poorly establish and evaluate this alignment, leading to undesired system behaviour and less effective evaluation.
For instance, the dataset in~\cite{zhou2022audio} shows a ``commonsense'' bias because it assumes that \emph{certain objects are always the sound source in some scenarios}.
Fig.~\ref{fig:motivation} shows an example of a scene from~\cite{zhou2022audio} with a bias toward the segmentation of the helicopter, even though other sound sources (e.g., person's speech or bird's singing) are plausible. Such biases can reduce cross-modal alignment understanding.
Another challenge in AVS datasets~\cite{zhou2022audio} is \emph{the localisation of the sound-producing object when multiple instances of the same object class are present}. 
While visual cues (e.g., motion or visual semantics) can help, some actions with silent audio can introduce false positives~\cite{morgado2021robust}. 
Spatial audio can also be helpful~\cite{wu2021binaural,rachavarapu2021localize}, as shown in Fig.~\ref{fig:process-stereo}, where we are more likely to segment the dog on the right if the spatial audio suggests that the sounding object is located on the right-hand side of the image. 
Regrettably, addressing the AVS dataset challenges mentioned above by collecting new datasets can be exorbitantly expensive. 
Therefore, we consider alternative data collection procedures to mitigate the problems described above and to effectively enable the training and evaluation of AVS methods that can better generalise to diverse AVS conditions.

Many AVL~\cite{chen2021localizing,mo2022localizing} and AVS~\cite{mao2023contrastive} methods rely on audio-visual contrastive learning (AV-CL). 
AV-CL bears some similarities with metric learning, particularly with respect to the selection of informative samples (also called hard sampling) for improving training efficiency and model performance~\cite{schroff2015facenet, oh2016deep, zheng2019hardness}.
In AVC tasks, such a selection of informative training samples is not well-studied, but it is critical because the more representative visual data can 
overpower the weaker audio modality~\cite{peng2022balanced}, resulting in false detections that are relatively independent of the audio-visual input, as shown in Fig.~\ref{fig:motivation} for TPAVI (columns 2 to 4).
Also, current AV-CL methods~\cite{mao2023contrastive,mo2022closer} consist of unsupervised learning approaches that treat each audio-visual data pair as an independent contrastive class.
However, such instance-based CL is unable to effectively mine informative samples that can mitigate the false detections mentioned above.

In this paper, we introduce a new cost-effective AVS data collection procedure for training and evaluating AVS methods that aim to mitigate the aforementioned problems of AVS datasets~\cite{zhou2022audio}, and a new AVS method developed to address the shortcomings of AV-CL approaches.
Our new AVS dataset collection and annotation, called Visual Post-production (VPO), consists of matching images from COCO~\cite{lin2014microsoft} and audio files from VGGSound~\cite{chen2020vggsound} based on the semantic classes of the visual objects of the images.
The proposed VPO dataset has three settings: 
1) the single sound-source (VPO-SS), which contains multiple visual objects, but just a single sounding object;  
2) multiple sound-source (VPO-MS), which  contains multiple visual objects with multiple sounding objects from different classes; and
3) multiple sound-source multi-instance (VPO-MSMI), which has multiple sets of visual objects from the same or different classes with multiple sounding objects. 
In these three settings, stereo sound is used to disambiguate visual objects.
We also propose a new AVS method, named contrastive audio-visual pairing (CAVP), with a supervised contrastive learning approach that leverages audio-visual pairings to mine informative contrastive samples. 
To summarise, our main contributions are
\begin{itemize}
    \item A new cost-effective strategy to build AVS datasets, named Visual Post-production (VPO), which aims to reduce the biases observed in current datasets~\cite{zhou2022audio} by pairing  images~\cite{lin2014microsoft} and audio files~\cite{chen2020vggsound} based on the visual classes of the image objects.
    Three new VPO benchmarks are built using this strategy: the single sound source (VPO-SS: single sounding object per image), multiple sound sources (VPO-MS: multiple sounding objects per image from separate classes), and multiple sound sources multi-instance (VPO-MSMI: multiple sets of sounding objects from the same class). 
    \item A new supervised audio-visual contrastive learning method that mines informative contrastive pairs from arbitrary audio-visual pairings to better constrain the learning of audio-visual embeddings.
    \item A thorough evaluation of SOTA AVS methods on AVSBench and VPO datasets. The methods are also assessed on AVS salient and semantically labelled objects with the resized image and traditional full-resolution setups.
\end{itemize}
We first show the effectiveness of our VPO strategy by modifying the AVSBench dataset~\cite{zhou2022audio} with the matching of AVSBench images with new VGGSound~\cite{chen2020vggsound} audio files from the same classes. We then train TPAVI~\cite{zhou2022audio} on the original and modified AVSBench datasets and show that both datasets lead to the equivalent performance of TPAVI on the testing set of AVSBench.
We also conducted experiments to test the segmentation accuracy of our proposed AVS method on AVSBench-Objects~\cite{zhou2022audio}, AVSBench-Semantics~\cite{zhou2023audio}, VPO-SS, VPO-MS and VPO-MSMI, and results display a consistent improvement of our method compared to the SOTA.

\section{Related Works}
\label{sec:related_works}

\noindent \textbf{Audio-visual Localisation (AVL)} is a binary classification task for detecting sounding visual objects in videos, using image sequences and audio signals. It employs unsupervised AVL training with ImageNet pre-trained backbone models~\cite{deng2009imagenet}. 
Prior research focused on creating joint audio-visual representations, with feature concatenation~\cite{arandjelovic2017look} or attention modules~\cite{chen2021localizing, mo2022closer, mo2022localizing}.
However, neglecting the contribution from audio can be a concern when audio and visual representations are not properly constrained~\cite{senocak2023sound}. This issue is addressed with contrastive learning~\cite{he2020momentum, chen2020simple, chen2020improved}, which emphasizes discriminative audio-visual feature learning for each instance~\cite{senocak2018learning, chen2021localizing, hu2022mix, mo2022closer, mo2022localizing, sun2023learning}.
However, Senocak et al.~\cite{senocak2023sound} argue that AVL with instance discrimination may hinder genuine cross-modal semantic understanding~\cite{morgado2021audio}. Hence, they propose leveraging a richer positive set, obtained through strong augmentation or nearest neighbours, to facilitate cross-modal alignment.
Despite these advancements, the lack of pixel-level annotation with semantic labels still hinders the accurate detection of visual objects.

\begin{figure*}[t!]
    \centering
    \includegraphics[width=.98\linewidth]{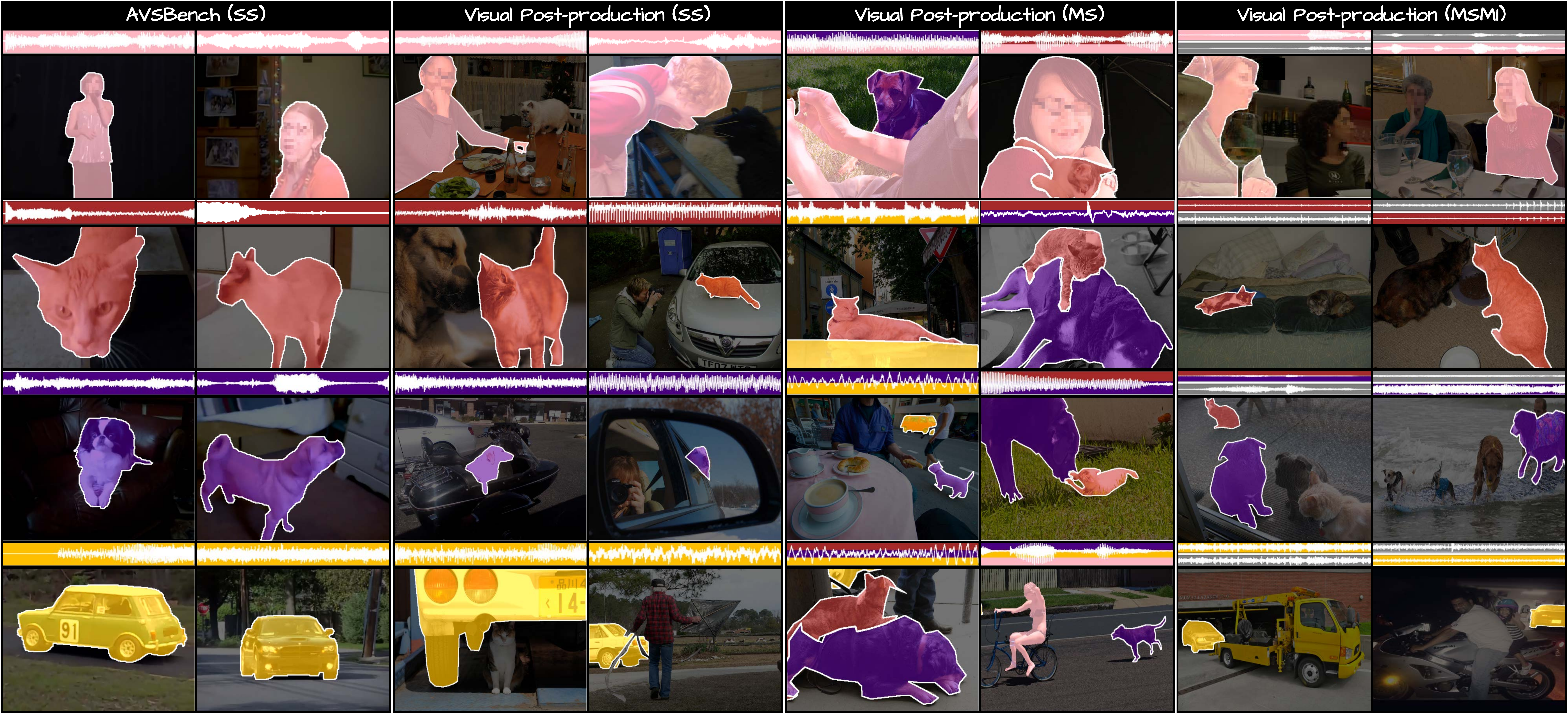}
    \vspace{-8pt}
    \caption{\textbf{VPO Benchmarks.} Using four classes, including ``\textcolor{mypink}{\textbf{female}}'', ``\textcolor{myred}{\textbf{cat}}'', ``\textcolor{mypurple}{\textbf{dog}}'', and ``\textcolor{myyellow}{\textbf{car}}'', the AVSBench (SS) (1st frame) provides pixel-level multi-class annotations to the images containing a single sounding object.  The proposed VPO benchmarks (2nd frame to 4th frame) pair a subset of the segmented objects in an image with relevant audio files to produce pixel-level multi-class annotations. 
    }
    \vspace{-10pt}
    \label{fig:dataset}
\end{figure*}

\noindent \textbf{Audio-visual Segmentation (AVS)} addresses the limitations observed in AVL by providing pixel-level binary annotations. Zhou et al.~\cite{zhou2022audio} introduced the AVSBench-Object and AVSBench-Semantic benchmarks~\cite{zhou2023audio}, which encompass single-source and multi-source AVS tasks for salient/multi-class object segmentation. The manual annotation of AVS datasets is costly and hence limits dataset diversity. Our VPO aims to mitigate this issue with off-the-shelf semantic segmentation datasets~\cite{lin2014microsoft} combined with audio data from YouTube~\cite{chen2020vggsound} to build datasets that facilitate model training and cross-modal alignment evaluations.

A recently published paper, developed in parallel with ours, proposes the AVS-Synthetic~\cite{liu2023annotation} benchmark aimed to reduce annotation costs by matching audio and visual categories to create a synthetic dataset. While our VPO and AVS-Synthetic share a similar concept of dataset creation, the data selection and partition process in AVS-Synthetic~\cite{liu2023annotation} has several problems, such as: 1) most annotated images consist of trivial cases containing a single sounding object, 2) it cannot assess models under different scenarios like single-source and multi-source settings, as proposed in AVSBench~\cite{zhou2022audio}, 3) the use of binary labels limits semantic understanding, and 4) it shows ambiguous cases where multiple instances (MI) of the same class are associated with single audio.
Our dataset collection addresses these four points, and in particular, it provides a cost-effective method to create spatial audio based on an object's relative position within the scene.

Like AVL,  AVS methods~\cite{liu2023audio,huang2023discovering, gao2023avsegformer,li2023catr,liu2023annotation,liu2023bavs,liu2023audiovisual} use audio for segmentation queries. For example, some methods adopt MaskFormer~\cite{cheng2021per} to perform image segmentation using audio queries and cross-attention layers. These methods benefit from the attention mechanism's ability to capture long-range dependencies and segment images, enhancing spatial-temporal reasoning~\cite{li2023catr} and task-related features~\cite{li2023catr,liu2023audiovisual,liu2023bavs,gao2023avsegformer,zhou2022audio, mao2023multimodal}. 
Also, certain methods~\cite{mao2023contrastive,mao2023multimodal} have investigated the utilization of conditional generative models~\cite{sohn2015learning,ho2020denoising}, alongside contrastive learning, to create discriminative latent spaces.
However, the reliance on binary annotations and image resizing limits their application and segmentation accuracy.

\noindent \textbf{Contrastive learning} has shown promise in AVL methods~\cite{chen2021localizing,hu2022mix,mo2022localizing,mo2022closer}. These methods bring together augmented representations from the same instance as positives while separating representation pairs from different instances as negatives within a batch. 
The issue with current AVL contrastive learning is its reliance on self-supervision~\cite{chen2020simple} to connect audio and visual representations of the same class. In our work, we propose a new supervised contrastive learning~\cite{khosla2020supervised, wang2021exploring, li2022targeted} that mines informative contrastive pairs from arbitrary audio-visual pairings to constrain the learning of audio-visual embeddings.

\section{Visual Post-production (VPO) Benchmark} \label{sec:vpo}

Our VPO benchmark includes three evaluation scenarios: 1) the single-source (VPO-SS) shown in Fig.~\ref{fig:dataset} (2nd frame), 2) the multi-source (VPO-MS) displayed in Fig.~\ref{fig:dataset} (3rd frame), and 3) the multi-source multi-instance (VPO-MSMI) displayed in Fig.~\ref{fig:dataset} (4th frame). 
VPO is built by combining images and semantic segmentation masks from COCO~\cite{lin2014microsoft} with audio files for the 21 COCO classes, including humans, animals, vehicles, sports, and electronics, sourced from VGGSound~\cite{chen2020vggsound,zhou2022audio}. The audio files were obtained from YouTube videos under the \textit{Creative Commons} license, each trimmed to 10 seconds, as verified by~\cite{chen2020vggsound}. 
We then randomly matched the COCO semantic segmentation masks with related audio files based on instance labels to create the VPO dataset. Below, we describe the three VPO settings: VPO-SS, VPO-MS, and VPO-MSMI.
For dataset statistics, examples and a detailed description of the collection process please refer to \textit{Supplementary Material}.
%

%

The \textbf{VPO-SS} comprises 12,202 samples (11,312 training and 890 testing samples), where a sample consists of an image, a pixel-level semantic segmentation mask of a single-sounding object, and an audio file of the sounding object class. 
During image collection, our process prioritized the inclusion of sounding visual classes from multiple categories within the same image, even when only one class matched the audio file. This strategy aims to reduce the ``commonsense'' bias in AVS datasets and avoid the dominance of a single visual object in the segmentation process, minimizing incidental correlations. 
This is done by ensuring images containing visual objects from various COCO classes are given higher priority than single-class images.
By employing this collection protocol, we aim to produce a benchmark that has a diverse set of sounding visual objects because we can match images with many different types of audio files that represent visual objects, resulting in numerous combinations, as demonstrated in Fig.~\ref{fig:dataset} (2nd frame).

The \textbf{VPO-MS} comprises 9,817 images, with 8,380 images for training and 1,437 for testing. Each image can include up to five sounding objects from the 21 COCO classes, where each visual object is accompanied by its pixel-level semantic segmentation mask and corresponding audio file. In total, the dataset contains 13,496 semantic segmentation masks.
Following the same VPO-SS strategy, we prioritise the collection of images that have sounding objects from multiple classes but exclude images that contain multiple instances of the same class.
Additionally, to prevent methods that detect all visual objects from an image from performing well in our benchmark, we randomly remove the sound of some visual objects from images containing more than two objects. 
We merge audio files from multiple sounding visual objects into a single file using addition operations performed on the waveform data~\cite{yang2022torchaudio,hu2022mix}. Fig.~\ref{fig:dataset} (3rd frame) shows VPO-MS examples.

\begin{figure}[t]
    \centering
    \includegraphics[width=0.98\linewidth]{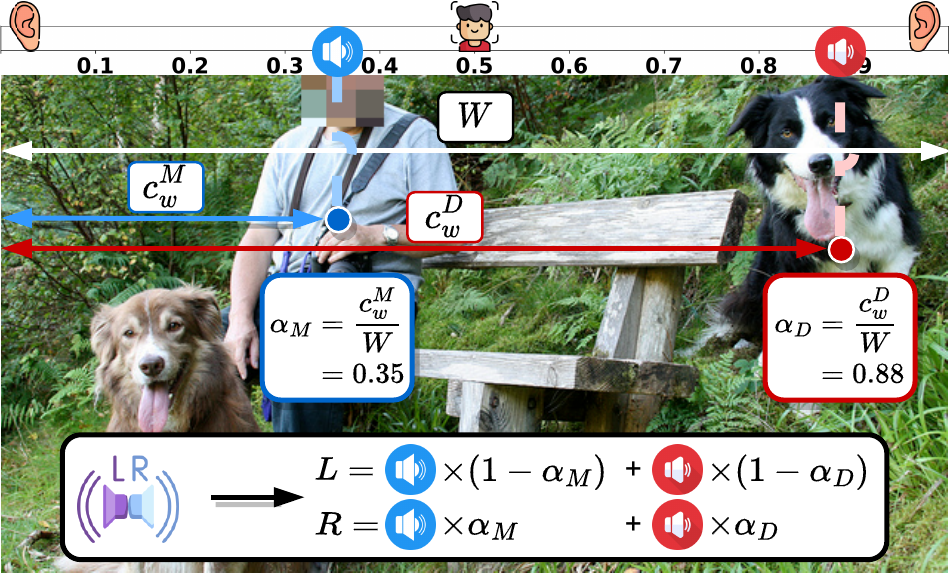}
    \vspace{-8pt}
    \caption{Synthesising stereo sound for the VPO-MSMI setting.}
    \vspace{-15pt}
    \label{fig:process-stereo}
\end{figure}

Based on VPO-MS, we additionally searched 3,038 images, each containing multiple instances that share a common semantic class. The resulting \textbf{VPO-MSMI}  contains 12,855 images, with 11,080 images for training and 1,775 for testing. The VPO-MSMI is a challenging AVS task, represented by the accurate localisation of sound sources amid multiple sources of sound with the assistance of spatial cues.
The main cue used to allow the segmentation of multiple instances in \textbf{VPO-MSMI} is the use of spatial audio to localize the sound source in an image. We simulate spatial audio with stereo sound by leveraging the object's spatial location information to modulate the volume of the left or right audio channel. Assuming we have an RGB image with resolution $H \times W$ and a particular object centred at $(c_h, c_w)$, then the relative position of that object w.r.t the width of the image is denoted by $\alpha_{i} = \frac{c_w}{W}$. For instance, we show an example of processing the audio files from a human and dog in Fig.~\ref{fig:process-stereo}, 
utilizing position coefficients ${\alpha_{M},\alpha_{D}}$ for volume control and deriving $c_w$ from ground-truth mask center of mass.
Based on this modelling method, we can work with an arbitrary number of sound sources.

\begin{figure*}[t]
    \centering
    \includegraphics[width=1\linewidth]{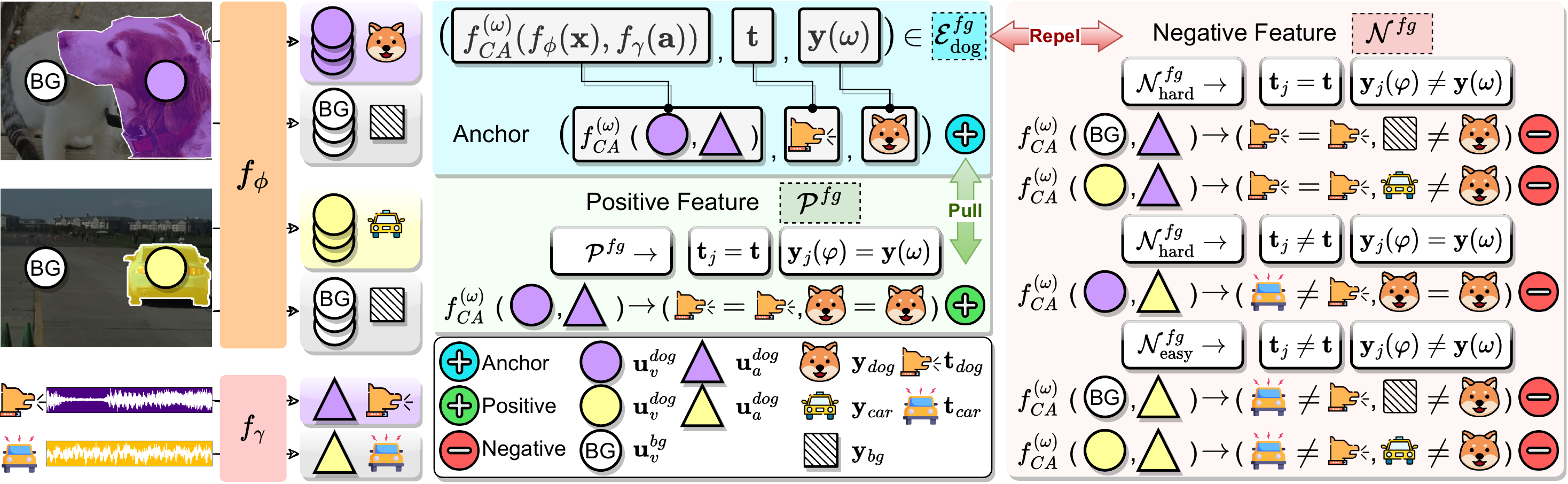}
    \vspace{-15pt}
    \caption{
    Illustration of our CAVP method for the ``Dog'' anchor. 
    Starting with the audio-visual foreground anchor set $\mathcal{E}^{\text{fg}}_{\text{dog}}$, we create the positive and negative audio-visual features denoted by $\mathcal{P}^{\text{fg}}$ and $\mathcal{N}^{\text{fg}} = \mathcal{N}^{\text{fg}}_{\mathsf{hard}} \cup \mathcal{N}^{\text{fg}}_{\mathsf{easy}}$ respectively defined in Eq.~\eqref{eq:positive_negative_sets}.
    The CAVP loss in Eq.~\eqref{eq:contrastive_loss_cavp} pulls the anchor and positive audio-visual features closer while repelling the anchor and negative audio-visual features. 
    } 
    \vspace{-15pt}
    \label{fig:avs-framework}
\end{figure*}

\section{Method}
\label{sec:method}

The multi-class audio-visual dataset is denoted as $\mathcal{D}=\{(\mathbf{a}_i, \mathbf{x}_i, \mathbf{y}_i, \mathbf{t}_i)\}_{i=1}^{|\mathcal{D}|}$, where $\mathbf{x}_i\in\mathcal{X}\subset \mathbb{R}^{H \times W \times 3}$ is an RGB image with resolution $H \times W$, $\mathbf{a}\in\mathcal{A}\subset \mathbb{R}^{T \times F}$ denotes the Mel Spectrogram audio representation with time $T$ and $F$ Mel filter banks,
$\mathbf{y}_i\in\mathcal{Y}\subset \{0, 1\}^{H \times W \times C}$ denotes the pixel-level ground truth for the $C$ classes (the background class is included in these $C$ classes), and $\mathbf{t}_i\in\mathcal{Y}\subset \{0, 1\}^{|C|}$ is a multi-label ground truth audio annotation.

\subsection{Preliminaries about Cross-Attention}
\label{sec:preliminaries}

Our goal is to learn the parameters $\theta \in \Theta$ for the model $f_{\theta}:\mathcal{X} \times \mathcal{A} \to [0,1]^{H \times W \times C}$, which comprises the image and audio backbones that extract features with $\mathbf{u}_a=f_\gamma(\mathbf{a})$ and $\mathbf{u}_v=f_\phi(\mathbf{x})$, respectively, where $\gamma,\phi \in \theta$, and $\mathbf{u}_a,\mathbf{u}_v \in \mathcal{U}$, with $\mathcal{U}$ denoting a unified feature space.
Our approach is similar to other early fusion methods~\cite{nagrani2021attention,zhou2022audio,gao2023avsegformer} that combine audio and video features with a multi-head attention block~\cite{vaswani2017attention} which estimates the co-occurrence of audio and visual data with $f_{MHA}(\mathbf{Q},\mathbf{K},\mathbf{V}) = \mathrm{softmax}\left(\frac{\mathbf{Q}\mathbf{K}^T}{\sqrt{D}}\right)\mathbf{V},$
where $\sqrt{D}$ denotes the scaling factor~\cite{vaswani2017attention}, and $\mathbf{Q},\mathbf{K},\mathbf{V}$ represent the query, key and value inputs.
Previous AVS methods~\cite{zhou2022audio, gao2023avsegformer} usually use audio as the cross-attention (CA) query $\mathbf{Q}$ to produce $\hat{\mathbf{u}}_{v} = \mathbf{u}_{v} \oplus \mathbf{u}_{v}\odot f_{MHA}(\mathbf{u}_{a},\mathbf{u}_{v},\mathbf{u}_{v})$, where $\oplus$ is the element-wise addition operator, and $\odot$ is the element-wise multiplication operator.
However, we empirically observe that the use of audio as query 
leads to a situation where the visual feature dominates the CA process, suppressing the audio representation.
We address this issue by using the visual feature as the query and removing the $\odot$ operation to produce  $\hat{\mathbf{u}}_{v} = f_{CA}(\mathbf{u}_{v},\mathbf{u}_{a})$, where $f_{CA}(\mathbf{u}_{v},\mathbf{u}_{a}) = \mathbf{u}_{v} \oplus f_{MHA}(\mathbf{u}_{v},\mathbf{u}_{a},\mathbf{u}_{a})$. 
Another important point about $f_{MHA}(.)$ is that we replace $\mathrm{softmax}(.)$ by $\mathrm{sigmoid}(.)$ to enable the highlighting of multiple regions of varying sizes in the attention map that is related to the audio. Please refer to the \textit{Supplementary Material} for the visual results.

\begin{table*}[!ht]
    \centering
    \caption{Quantitative ($\mathcal{J}$, $\mathcal{F}$) audio-visual segmentation results (in \%) on AVSBench dataset~\cite{zhou2022audio,zhou2023audio} (resized to 224$\times$224) with ResNet50~\cite{he2016deep} backbone. Best results in \textbf{bold}, second best \underline{underlined}. Improvements against the second best are in the brackets.
    }
    \vspace{-5pt}
    \label{tab:avsbench_224_r50}
    \resizebox{0.98\linewidth}{!}{
    \begin{tabular}{
    !{\vrule width 1.5pt}c
    !{\vrule width 1.5pt}c
    !{\vrule width 1.5pt}
    ccc|ccc|ccc!{\vrule width 1.5pt}
    } 
\specialrule{1.5pt}{0pt}{0pt}
\multicolumn{11}{!{\vrule width 1.5pt}c!{\vrule width 1.5pt}}{ImageNet Pretrained Backbone} \\
\specialrule{1.5pt}{0pt}{0pt}
\multirow{2}{*}{D-ResNet50~\cite{he2016deep}} & 
\multirow{2}{*}{Method} & 
\multicolumn{3}{c|}{AVSBench-Object (SS)} & 
\multicolumn{3}{c|}{AVSBench-Object (MS)} & 
\multicolumn{3}{c!{\vrule width 1.5pt}}{AVSBench-Semantics} \\
\cline{3-11}
\rule{0pt}{10pt}
~ & ~ &  
$\mathcal{J}$\&$\mathcal{F}$ $\uparrow$ & $\mathcal{J}$ $\uparrow$ & $\mathcal{F}$ $\uparrow$ & 
$\mathcal{J}$\&$\mathcal{F}$ $\uparrow$ & $\mathcal{J}$ $\uparrow$ & $\mathcal{F}$ $\uparrow$ &  
$\mathcal{J}$\&$\mathcal{F}$ $\uparrow$ & $\mathcal{J}$ $\uparrow$ & $\mathcal{F}$ $\uparrow$ \\ 
\specialrule{1.5pt}{0pt}{0pt}
\multirow{5}{*}{Transformer} 
& CATR~\cite{li2023catr} & 80.70 & 74.80 & 86.60 & \underline{59.05} & \underline{52.80} & \underline{65.30} & - & - & - \\ 
& AuTR~\cite{liu2023audio} & 80.10 & 75.00 & 85.20 & 55.30 & 49.40 & 61.20 & - & - & - \\ 
& AVSegFormer~\cite{gao2023avsegformer} & 80.67 & 76.54 & 84.80 & 56.17 & 49.53 & 62.80 & 27.12 & \underline{24.93} & 29.30 \\ 
& AVSC~\cite{liu2023audiovisual} & 81.13 & 77.02 & 85.24 & 55.55 & 49.58 & 61.51 & - & - & - \\
& BAVS~\cite{liu2023bavs} & \underline{81.63} & \underline{77.96} & 85.29 & 56.30 & 50.23 & 62.37 & \underline{27.16} & 24.68 & \underline{29.63} \\ 
\hline
\multirow{4}{*}{\makecell{Per-pixel\\Classification}} 
& TPAVI~\cite{zhou2022audio} & 78.80 & 72.79 & 84.80 & 52.84 & 47.88 & 57.80 & 22.69 & 20.18 & 25.20 \\ 
& AVSBG~\cite{hao2023improving} & 79.77 & 74.13 & 85.40 & 50.88 & 44.95 & 56.80 & - & - & - \\ 
& ECMVAE~\cite{mao2023multimodal} & 81.42 & 76.33 & 86.50 & 54.70 & 48.69 & 60.70 & - & - & - \\ 
& DiffusionAVS~\cite{mao2023contrastive} & 81.35 & 75.80 & \underline{86.90} & 55.94 & 49.77 & 62.10 & - & - & - \\ 
\rowcolor{LightCyan}\cellcolor{white}
& \textbf{Ours} & 
\textbf{83.84} & \textbf{78.78} & \textbf{88.89} & 
\textbf{61.48} & \textbf{55.82} & \textbf{67.14} & 
\textbf{32.83} & \textbf{30.37} & \textbf{35.29} \\ \hline
\specialrule{1.5pt}{0pt}{0pt}
\multicolumn{11}{!{\vrule width 1.5pt}c!{\vrule width 1.5pt}}{COCO Pretrained Backbone} \\
\specialrule{1.5pt}{0pt}{0pt}
Transformer & AQFormer & 81.70 & 77.00 & 86.40 & 61.30 & 55.70 & 66.90 & - & - & - \\ 
\makecell{Per-pixel Cls.}& 
\cellcolor{LightCyan}\textbf{Ours} & \cellcolor{LightCyan}\textbf{83.75} & \cellcolor{LightCyan}\textbf{78.72} & \cellcolor{LightCyan}\textbf{88.77} & \cellcolor{LightCyan}\textbf{62.34} & \cellcolor{LightCyan}\textbf{56.42} & \cellcolor{LightCyan}\textbf{68.25} & - & - & - \\ 
\specialrule{1.5pt}{0pt}{0pt}
\end{tabular}
}
\vspace{-10pt}
    
\end{table*}

\vspace{-10pt}
\subsection{Contrastive Audio-Visual Pairing (CAVP)}
\label{sec:CAVP}

The ultimate goal of our contrastive learning is to discriminate the positive fusion of feature representations from the same semantic class and the negative fusion of feature representations from different semantic classes.
Previous works~\cite{khosla2020supervised,liu2021contrastive} suggest that the discrimination between samples is critical for contrastive learning, so simply drawing random negative samples from original audio-visual pairs could limit the representation learning effectiveness.
A practical way to create an informative contrastive dataset is to mine all possible combinations of audio-visual pairs in the latent space.
However, it is not possible to create some of the negative pairs because we do not have access to all instance segmentation masks for each image (e.g., we cannot create $f_{CA}(\mathbf{u}_v^{cat},\mathbf{u}_a^{dog})$ in Fig.~\ref{fig:avs-framework} because we do not have pixels labelled as ``cat'').
Motivated by our VPO procedure in Sec.~\ref{sec:vpo}, we observe that this barrier can be overcome 
by randomly paring audio and visual features within the batch, as shown in Fig.~\ref{fig:avs-framework},
offering us the potential to mine most of the conceivable audio-visual combinations.

CAVP starts by dividing the original audio-visual dataset $\mathcal{D}$ into a visual dataset $\mathcal{D}^v=\{(\mathbf{x}_i, \mathbf{y}_i)\}_{i=1}^{|\mathcal{D}|}$ and audio dataset $\mathcal{D}^a=\{(\mathbf{a}_{k}, \mathbf{t}_{k})\}_{k=\mathsf{perm}(\{1, ...,|\mathcal{D}|\})}$, where $j$ is a permutation of the index set of $\mathcal{D}$. 
We define the randomly paired audio-visual feature dataset as:
\begin{equation}
\scalebox{0.8}{$
    \mathcal{Z}  = \left \{\left(\mathbf{z},\mathbf{t},\mathbf{y}(\omega)\right)|\mathbf{z}=f^{(\omega)}_{CA}(f_\phi(\mathbf{x}),f_\gamma(\mathbf{a})), (\mathbf{a},\mathbf{t})\in\mathcal{D}^a, (\mathbf{x},\mathbf{y})\in\mathcal{D}^v \right \},
    $}
    \label{eq:rand_pair_AV_data}
\end{equation}
where $\omega \in \Omega$ is the lattice of size $H \times W$, and $f^{(\omega)}_{CA}(.)$ is the cross-attention output at lattice position $\omega$. 
To simplify the notation, we represent $\mathbf{v}(\omega)=(\mathbf{z},\mathbf{t},\mathbf{y}(\omega))$ and define the audio-visual anchor sets as: 
\begin{equation}
\scalebox{0.85}{$
\begin{aligned}
    \mathcal{E}^{\text{fg}} &= \big \{\mathbf{v}(\omega)|  \mathbf{v}(\omega) \in \mathcal{Z}, (\mathbf{t} = \mathbf{y}(\omega)) \wedge (\mathbf{t}\ne\text{bg}) \wedge (\mathbf{y}(\omega)\ne\text{bg}) \big \}, \\
    \mathcal{E}^{\text{unknow}} &= \big \{\mathbf{v}(\omega)|  \mathbf{v}(\omega) \in \mathcal{Z}, (\mathbf{t}_j \ne \mathbf{y}_j(\omega)) \wedge (\mathbf{y}_j(\omega) = \text{bg}) \big \}, \\
    \mathcal{E}^{\text{bg}} &= \mathcal{Z} \setminus (\mathcal{E}^{\text{fg}} \cup \mathcal{E}^{\text{unknow}}),
\end{aligned}
$}
\label{eq:anchor_sets}
\end{equation}
where $\wedge$ and $\vee$ are respectively the ``AND'' and ``OR'' logic operators, and ``bg'' is the background class label. 
The foreground anchor set $\mathcal{E}^{\text{fg}}$ contains samples from $\mathcal{Z}$ in Eq.~\eqref{eq:rand_pair_AV_data} that have the same audio and visual labels, and both are different from the background class; the $\mathcal{E}^{\text{unknow}}$ represents the anchors with uncertain semantic meaning due to the lack of instance segmentation masks for all the available targets, and the background anchor set $\mathcal{E}^{\text{bg}}$ are all the samples in $\mathcal{Z}$ that are not in $\mathcal{E}^{\text{fg}} \cup \mathcal{E}^{\text{unknow}}$.

The mining of informative contrastive samples explores all combinations of audio-visual features after cross-attention fusion to form positive sets that will enable the enhancement of the similarity between semantically related samples while diminishing the similarity of samples with different semantic concepts.
For foreground anchors in $\mathcal{E}^{\text{fg}}$, its contrastive positive set $\mathcal{P}^{\text{fg}}$ and negative set $\mathcal{N}^{\text{fg}} = \mathcal{N}^{\text{fg}}_{\mathsf{hard}} \bigcup \mathcal{N}^{\text{fg}}_{\mathsf{easy}}$ are defined by (see Fig.~\ref{fig:avs-framework}): 
\begin{equation}
\scalebox{0.92}{$
\begin{aligned}
\begin{alignedat}{2}
    \mathcal{P}^{\text{fg}}(\mathbf{v}(\omega)) &= \big \{ \mathbf{z}_j | \mathbf{v}_j(\varphi)\in\mathcal{Z},&&((\mathbf{t}_j = \mathbf{t}) \wedge (\mathbf{y}_j(\varphi) = \mathbf{y}(\omega))) \big \},\\ 
    \mathcal{N}^{\text{fg}}_{\mathsf{hard}}(\mathbf{v}(\omega)) &= \big \{ \mathbf{z}_j |  \mathbf{v}_j(\varphi)\in\mathcal{Z},&&((\mathbf{t}_j = \mathbf{t}) \wedge (\mathbf{y}_j(\varphi) \ne \mathbf{y}(\omega))) \\ 
                                            &\hspace{2.5cm}\vee && ((\mathbf{t}_j \ne \mathbf{t}) \wedge (\mathbf{y}_j(\varphi) = \mathbf{y}( \omega))) \big \}, \\
    \mathcal{N}^{\text{fg}}_{\mathsf{easy}}(\mathbf{v}(\omega)) & = \big \{ \mathbf{z}_j | \mathbf{v}_j(\varphi)\in\mathcal{Z},&&((\mathbf{t}_j \neq \mathbf{t}) \wedge (\mathbf{y}_j(\varphi) \neq \mathbf{y}(\omega))) \big \},
\end{alignedat}
\end{aligned}
$}
\label{eq:positive_negative_sets}
\end{equation}
where $\varphi \in \Omega $. For background cases in $\mathcal{E}^{\text{bg}}$, the contrastive positive set is $\mathcal{P}^{\text{bg}}= \mathcal{E}^{\text{bg}}$, while the negative set is $\mathcal{N}^{\text{bg}}=\mathcal{E}^{\text{fg}}$.
Let us represent the set of anchors by $\mathcal{E} = \mathcal{E}^{\text{fg}} \bigcup \mathcal{E}^{\text{bg}}$, the set of positives with $\mathcal{P}$ that is equal to $\mathcal{P}^{\text{fg}}$ if the anchor is from $\mathcal{E}^{\text{fg}}$, or equal to $\mathcal{P}^{\text{bg}}$ if the anchor is from $\mathcal{E}^{\text{bg}}$ (and similarly for $\mathcal{N}$ w.r.t. $\mathcal{N}^{\text{fg}}$ or $\mathcal{N}^{\text{bg}}$).
Adopting the supervised InfoNCE~\cite{khosla2020supervised} as the objective function 
to pull the anchor $\mathbf{v}(\omega) \in \mathcal{E}$ and respective positive audio-visual features closer while repelling anchors and their negative audio-visual features, 
we define the following loss:
\begin{equation}
\begin{split}
    \ell_{\text{CP}}&(\mathbf{v}(\omega)) = \frac{1}{|\mathcal{P}(\mathbf{v}(\omega))|}\sum_{\mathbf{z}_p \in\mathcal{P}(\mathbf{v}(\omega))} \\ 
    &-\log\frac{\exp{(\mathbf{z}\cdot \mathbf{z}_p/\tau)}}{\exp{(\mathbf{z}\cdot \mathbf{z}_p/\tau)}+\sum_{\mathbf{z}_n\in\mathcal{N}(\mathbf{v}(\omega))}\exp{(\mathbf{z}\cdot \mathbf{z}_n/\tau)}},
\end{split}
\label{eq:contrastive_loss_cavp}
\end{equation}
where $\mathbf{z}$ is the anchor feature from $\mathbf{v}(\omega)$, 
and $\tau$ is the temperature hyper-parameter. 

The overall training loss function is defined as:
\begin{equation}
\scalebox{0.78}{$
\begin{aligned}
    \ell(\mathcal{D},\theta) =   \frac{1}{|\mathcal{D}| |\Omega|}  \sum_{i=1}^{|\mathcal{D}| } 
     &\sum_{\varphi\in \Omega} \Big( \ell_{\text{CE}}(\mathbf{y}_i(\varphi),\mathbf{\hat{y}}_i(\varphi))\Big ) + \frac{1}{|\mathcal{E}|} \sum_{\mathbf{v}(\omega) \in \mathcal{E}} \Big( \ell_{\text{CP}}(\mathbf{v}(\omega)) \Big ),
\end{aligned}    
$}
\label{eq:main_loss}
\end{equation}
where $\ell_{\text{CE}}(.)$ is the cross-entropy loss, $\mathbf{\hat{y}}=f_{\theta}(\mathbf{x},\mathbf{a})$ is the model (parameterised by $\theta$) prediction, with $\hat{\mathbf{y}} \in [0,1]^{H \times W \times C}$, $\Omega$ is the image lattice, and 
$\ell_{\text{CP}}(.)$ is our contrastive loss, calculated based on the anchor sets $\mathcal{E}$ from Eq.\eqref{eq:anchor_sets}, as specified in Eq.~\eqref{eq:contrastive_loss_cavp}.

\begin{table*}[!ht]
    \centering
    \caption{Quantitative (mIoU, $F_{\beta})$ audio-visual segmentation results (in \%) on AVSBench dataset~\cite{zhou2022audio,zhou2023audio} (resized to 224$\times$224) with ResNet50~\cite{he2016deep} backbone. * indicate the our initial results. 
    }
    \label{tab:avsbench_224_r50_dataset_level}
    \begin{tabular}{!{\vrule width 1.5pt}c!{\vrule width 1.5pt}c!{\vrule width 1.5pt}cc|cc|cc!{\vrule width 1.5pt}} 
\specialrule{1.5pt}{0pt}{0pt}
\multirow{2}{*}{D-ResNet50~\cite{he2016deep}} & \multirow{2}{*}{Method} & \multicolumn{2}{c|}{AVSBench-Object (SS)} & \multicolumn{2}{c|}{AVSBench-Object (MS)} & \multicolumn{2}{c!{\vrule width 1.5pt}}{AVSBench-Semantics} \\
\cline{3-8}
\rule{0pt}{10pt}
~ & ~ &  mIoU $\uparrow$ & $F_{\beta}$ $\uparrow$ &  mIoU $\uparrow$ & $F_{\beta}$ $\uparrow$ &  mIoU $\uparrow$ & $F_{\beta}$ $\uparrow$  \\ 
\specialrule{1.5pt}{0pt}{0pt}
\multirow{2}{*}{\makecell{Per-Pixel\\Classification}} & CAVP* & 85.77 & 92.86 & 62.39 & 73.62 & 44.70 & 57.76 \\ 
~ & CAVP & 89.43 & 94.50 & 72.79 & 83.05 & 44.70 & 57.76 \\ 
\specialrule{1.5pt}{0pt}{0pt}
    \end{tabular}
    
\end{table*}
\begin{table*}[!ht]
    \centering
    \caption{Quantitative (mIoU, $F_{\beta}$, FDR) audio-visual segmentation results (in \%) on AVSBench-Semantic dataset~\cite{zhou2023audio} (original resolution) with ResNet50~\cite{he2016deep} backbone. 
    Best results in \textbf{bold}, second best \underline{underlined}. Improvements against the second best are in the last row.
    }
    \vspace{-5pt}
    \label{tab:avsbench_r50}
    \rowcolors{5}{LightCyan}{LightCyan}
    \def\arraystretch{1.1}
    \resizebox{0.98\linewidth}{!}{
    \begin{tabular}{!{\vrule width 1.5pt}c!{\vrule width 1.5pt}c!{\vrule width 1.5pt}ccc|ccc|ccc!{\vrule width 1.5pt}} 
\specialrule{1.5pt}{0pt}{0pt}
\multirow{2}{*}{D-ResNet50~\cite{he2016deep}} & \multirow{2}{*}{Method} & \multicolumn{3}{c|}{AVSBench-Semantics (SS)} & \multicolumn{3}{c|}{AVSBench-Semantics (MS)} & \multicolumn{3}{c!{\vrule width 1.5pt}}{AVSBench-Semantics} \\
\cline{3-11}
\rule{0pt}{10pt}
~ & ~ &  mIoU $\uparrow$ & $F_{\beta}$ $\uparrow$ & FDR $\downarrow$ & mIoU $\uparrow$ & $F_{\beta}$ $\uparrow$ & FDR $\downarrow$ & mIoU $\uparrow$ & $F_{\beta}$ $\uparrow$ & FDR $\downarrow$\\ 
\specialrule{1.5pt}{0pt}{0pt}
\global\let\CT@@do@color\relax 
\multirow{1}{*}{Transformer} 
& AVSegFormer~\cite{gao2023avsegformer} & \underline{68.95} & \underline{82.64} & 15.45 & 33.23 & 45.63 & 43.16 & 41.48 & 56.21 & 38.77 \\ \hline
\multirow{2}{*}{Per-pixel Classification} 
& TPAVI~\cite{zhou2022audio} & 64.30 & 81.06 & \underline{14.81} & \underline{36.29} & \underline{50.36} & \underline{40.61} & \underline{43.39} & \underline{59.24} & \underline{34.66} \\ 
& \global\let\CT@@do@color\oriCT@@do@color\textbf{Ours} & \textbf{73.08} & \textbf{85.57} & \textbf{12.64} & \textbf{46.40} & \textbf{60.25} & \textbf{32.11} & \textbf{50.75} & \textbf{64.57} & \textbf{32.25} \\ \hline 
Improvement & \textbf{Ours} & \improve{+4.13} & \improve{+2.93} & \drop{-2.17} & \improve{+10.11} & \improve{+9.89} & \drop{-8.50} & \improve{+7.36} & \improve{+5.33} & \drop{-2.41} \\
\specialrule{1.5pt}{0pt}{0pt}
    \end{tabular}
}
\vspace{-10pt}
    
\end{table*}
\begin{table*}[!ht]
    \centering
    \caption{Quantitative (mIoU, $F_{\beta}$, FDR) audio-visual segmentation results (in \%) on VPO dataset (original resolution) with ResNet50~\cite{he2016deep} backbone. Best results in \textbf{bold}, second best \underline{underlined}. Improvements against the second best are in the last row.}
    \label{tab:vpo_resnet50}
    \vspace{-5pt}
    \rowcolors{5}{LightCyan}{LightCyan}
    \resizebox{0.98\linewidth}{!}{
    \begin{tabular}{!{\vrule width 1.5pt}c!{\vrule width 1.5pt}c!{\vrule width 1.5pt}ccc|ccc|ccc!{\vrule width 1.5pt}} 
    \hline
\specialrule{1.5pt}{0pt}{0pt}
\multirow{2}{*}{D-ResNet50~\cite{he2016deep}} & \multirow{2}{*}{Method} & \multicolumn{3}{c|}{VPO (SS)} & \multicolumn{3}{c|}{VPO (MS)} & \multicolumn{3}{c!{\vrule width 1.5pt}}{VPO (MSMI)} \\
\cline{3-11}
\rule{0pt}{10pt}
~ & ~ &  mIoU $\uparrow$ & $F_{\beta}$ $\uparrow$ & FDR $\downarrow$ & mIoU $\uparrow$ & $F_{\beta}$ $\uparrow$ & FDR $\downarrow$ & mIoU $\uparrow$ & $F_{\beta}$ $\uparrow$ & FDR $\downarrow$\\ 
\specialrule{1.5pt}{0pt}{0pt}
\global\let\CT@@do@color\relax 
\multirow{1}{*}{Transformer} 
& AVSegFormer~\cite{gao2023avsegformer} & \underline{57.55} & \underline{73.03} & \underline{19.76} & \underline{58.33} & \underline{74.28} & \underline{22.13} & \underline{54.22} & \underline{70.39} & \underline{25.51} \\ \hline
\multirow{2}{*}{Per-pixel Classification} 
& TPAVI~\cite{zhou2022audio} & 52.75 & 69.54 & 22.83 & 54.30 & 71.95 & 22.45 & 51.73 & 68.85 & 26.75 \\ 
& \global\let\CT@@do@color\oriCT@@do@color\textbf{Ours} & \textbf{62.31} & \textbf{78.46} & \textbf{13.56} & \textbf{64.31} & \textbf{78.92} & \textbf{18.67} & \textbf{60.36} & \textbf{75.60} & \textbf{22.12} \\ \hline
Improvements & \textbf{Ours} & \improve{+4.76} & \improve{+5.43} & \drop{-6.20} & \improve{+5.98} & \improve{+4.64} & \drop{-3.46} & \improve{+6.14} & \improve{+5.21} & \drop{-3.39} \\
\specialrule{1.5pt}{0pt}{0pt}
    \end{tabular}
}
\vspace{-15pt}
\end{table*}
\begin{table}[t!]
\centering
\caption{Ablation study of the CAVP components.}
\vspace{-5pt}
\resizebox{0.98\linewidth}{!}{%
\begin{tabular}{
!{\vrule width 1.5pt}c!{\vrule width 1.5pt}P{50pt}|P{50pt}|P{50pt}|P{50pt}!{\vrule width 1.5pt}
} 
\specialrule{1.5pt}{0pt}{0pt}
\multirow{2}{*}{Method} & \multicolumn{2}{c|}{AVSBench-Semantics} & \multicolumn{2}{c!{\vrule width 1.5pt}}{VPO (MSMI)} \\
\cline{2-5}
\rule{0pt}{10pt}
~ &  mIoU $\uparrow$ & $F_{\beta}$ $\uparrow$ &  mIoU $\uparrow$ & $F_{\beta}$ $\uparrow$ \\
\specialrule{1.5pt}{0pt}{0pt}
TPAVI~\cite{zhou2022audio} & 42.74 & 58.11 & 52.83 & 70.39 \\ \hline
$f_{CA}(\cdot)$ & 47.18 & 61.74 & 58.50 & 73.19 \\ \hline
SupCon & 48.72 & 61.85 & 59.35 & 74.20 \\ \hline
CAVP & 50.75 & 62.31 & 64.31 & 64.13 \\ \hline

\specialrule{1.5pt}{0pt}{0pt}
\end{tabular}
\label{tab:ablation_components}
}
\vspace{-15pt}
\end{table}

\section{Experiments}

\subsection{Implementation Details}

\noindent \textbf{Evaluation protocols:}
We first adopt the widely used evaluation protocols for the AVSBench-Object~\cite{zhou2022audio} (including single-source (SS) and multi-source (MS) with binary labels) and AVSBench-Semantics dataset (with multi-class labels)~\cite{zhou2023audio} by resizing all images to 224 $\times$ 224 for fair comparison. 
Note that the use of binary labels and image resizing limits the application scope as well as the model performance. 
Hence, we follow traditional segmentation benchmarks~\cite{cordts2016cityscapes, everingham2015pascal, zhou2019semantic, lin2014microsoft, neuhold2017mapillary} and use original image resolution for training and testing.
Also following previous AVS methods~\cite{zhou2022audio,zhou2023audio}, we calculate mean intersection over union (mIoU)~\cite{everingham2015pascal} to quantify the average segmentation quality and use $F_{\beta}$ score with $\beta^2=0.3$~\cite{martin2004learning,zhou2022audio} to measure precision and recall performance and false detection rate (FDR) to highlight the false positive classification in a pixel-wise manner. For training and inference details, please refer to \textit{Supplementary Material}.


\subsection{Results}

\noindent \textbf{Performance on resized AVSBench.}\label{sec:result_avsbench_resize}
We divide the performance comparison on AVSBench into the AVSBench-Objects (SS \& MS)~\cite{zhou2022audio} and AVSBench-Semantics~\cite{zhou2023audio}. We compare the performance of SOTA methods with our CAVP in Tab.~\ref{tab:avsbench_224_r50} using Jaccrd index ($\mathcal{J}$) and F-Score ($\mathcal{F}$) based on~\cite{zhou2022audio}, which shows that our model surpasses the second-best methods w.r.t. mIoU by of $2.21\%$ on AVSBench-Object (SS)~\cite{zhou2022audio}, $2.43\%$ on AVSBench-Object (MS)~\cite{zhou2022audio} and $5.68\%$ on AVSBench-Semantics~\cite{zhou2023audio} using ResNet50~\cite{he2016deep} backbone. 
We separate the AQFormer~\cite{huang2023discovering} in the bottom of Tab.~\ref{tab:avsbench_224_r50} as their backbone model utilizes pre-trained weights based on COCO~\cite{lin2014microsoft} instead of ImageNet~\cite{deng2009imagenet}.
Additionally, we report model performance under conventional semantic segmentation evaluation metrics, same as Pascal VOC~\cite{everingham2015pascal} and CityScape~\cite{cordts2016cityscapes} as this method do not skip the no sounding frames during evaluation.

\begin{figure*}[!t]
    \centering
    \includegraphics[width=0.97\linewidth,height=6cm]{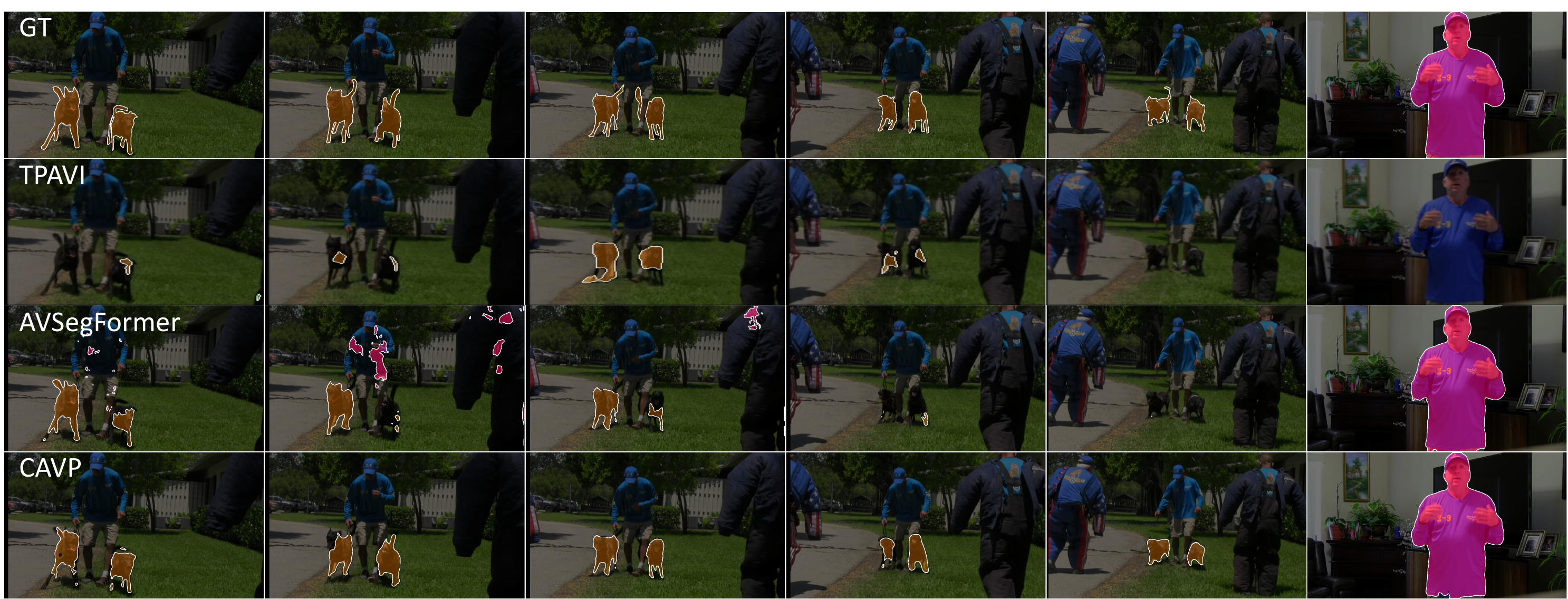}
    \vspace{-10pt}
    \caption{Qualitative audio-visual segmentation results on AVSBench-Semantics~\cite{zhou2023audio} by TPAVI~\cite{zhou2022audio}, AVSegFormer~\cite{gao2023avsegformer}, and our CAVP, which can be compared with the ground truth (GT) of the first row. 
    }
    \vspace{-10pt}
    \label{fig:avss-visual}
\end{figure*}

\noindent \textbf{Performance on original resolution AVSBench.} We introduce a new benchmark based on the AVSBench-Semantics dataset~\cite{zhou2023audio} in Tab.~\ref{tab:avsbench_r50} 
to address the absence of the original resolution images
in previous benchmarks shown in Tab.~\ref{tab:avsbench_224_r50}. 
In this new benchmark, we use AVSegFormer~\cite{gao2023avsegformer}\footnote{\url{https://github.com/vvvb-github/AVSegFormer}} and TPAVI~\cite{zhou2022audio}\footnote{\url{https://github.com/OpenNLPLab/AVSBench}} as the baseline methods\footnote{We cannot include other methods from Tab.~\ref{tab:avsbench_224_r50} as they were not publicly available online at the time of submission.}. To save training and evaluation time, we train on the entire training set of AVSBench-Semantics and test the model on AVSBench-Semantics (SS) and AVSBench-Semantics (MS) subsets, as well as the entire testing set for AVSBench-Semantics to show partitioned model performance.
The results on the entire AVSBench-Semantics in the last columns of Tables~\ref{tab:avsbench_224_r50} and~\ref{tab:avsbench_r50} show a significant mIoU improvement of $+16.55\%$ for AVSegFromer~\cite{gao2023avsegformer}, $+23.21\%$ for TPAVI, and +$6.05\%$ for our method when compared to the results obtained with low image resolution under ResNet-50 backbone~\cite{he2016deep}. 
These results suggest the importance of using the original resolution in the AVSBench-Semantics dataset~\cite{zhou2023audio}.
Also in Tab.~\ref{tab:avsbench_r50}, results show that our method consistently outperforms the baselines by a minimum of $4.13\%$ and $2.93\%$ improvements in mIoU and $F_{\beta}$, respectively. 
We show a visualisation of a 6-second video clip in Fig.~\ref{fig:avss-visual} that displays a qualitative comparison between TPAVI, AVSegFormer and our CAVP. 
Notice how our method approximates the ground truth segmentation of the sounding objects more consistently than the other methods.
Please refer to the \textit{Supplementary Material} for more qualitative results.

\begin{figure}[t]
    \centering
    \includegraphics[width=0.97\linewidth]{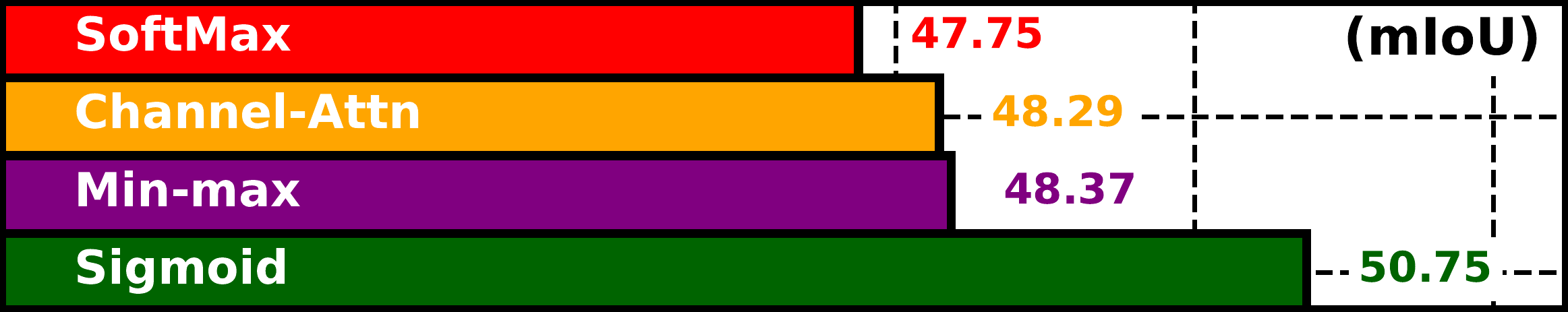}
    \vspace{-8pt}
    \caption{mIoU results for Activation functions used by $f_{CA}(.)$ on AVSBench-Semantic~\cite{zhou2023audio} with D-ResNet50~\cite{he2016deep} backbone.}
    \vspace{-15pt}
    \label{fig:ablation_ca}
\end{figure}

\noindent \textbf{Effectiveness of Visual Post-production (VPO):} To test the effectiveness of our VPO approach to build a benchmark dataset, we simulate VPO on the initial single source AVSBench-Object~\cite{zhou2022audio}. This involves the substitution of the original audio waveform data with samples from the same category taken from VGGSound~\cite{chen2020vggsound} (emulating the way we build the VPO datasets). We denote this dataset as AVS-M-SS. 
We directly use the original TPAVI code~\cite{zhou2022audio} and run the model training and testing three times for both the modified and the original datasets. 
On AVSBench-Object (SS), TPAVI achieves 
mIoU=$72.01\pm0.7\%$ and $F_{\beta}=83.36\pm1.2\%$ (mean $\pm$ standard deviation), and on AVS-M-SS, TPAVI achieves 
mIoU=$72.07\pm0.7\%$ and $F_{\beta}=83.70\pm1.0\%$. 
The p-values (two-sided t-test) of mIoU and $F_{\beta}$ are $0.66$ and $0.73$, respectively, which means that we fail to reject the null hypnosis that the performance on the VPO dataset is significantly different from the original one. 
This experiment shows the validity of building AVS datasets using pairs of audio-visual data obtained by
matching images and audio based on the semantic classes
of the visual objects of the images.

\noindent \textbf{Performance on VPO:}
We show model performance on our VPO benchmarks in Tab.~\ref{tab:vpo_resnet50} with ResNet50~\cite{he2016deep} image backbone. Our method outperforms the baseline methods on all experimental settings by a minimum of $4.76\%$ and $3.46\%$ for mIoU and false detection rate, respectively.


\begin{table}[!t]
\centering
\caption{Training with different proportions of positive and negative samples in~\eqref{eq:contrastive_loss_cavp}. 
We use D-ResNet50~\cite{he2016deep} and DeepLabV3+~\cite{chen2018encoder} as the backbone and the segmentation architecture, respectively.}
\vspace{-5pt}
\resizebox{0.98\linewidth}{!}{%
\begin{tabular}{!{\vrule width 1.5pt}c|c!{\vrule width 1.5pt}ccc!{\vrule width 1.5pt}} 
\specialrule{1.5pt}{0pt}{0pt}
\multicolumn{2}{!{\vrule width 1.5pt}c!{\vrule width 1.5pt}}{Proportion ($\%$)}   & \multicolumn{3}{c!{\vrule width 1.5pt}}{AVSBench-Semantics~\cite{zhou2023audio}} \\
\cline{1-5}
\rule{0pt}{10pt}
Pos Sample & Neg Sample & mIoU $\uparrow$ & $F_\beta$ $\uparrow$  &
FDR $\downarrow$ \\
\specialrule{1.5pt}{0pt}{0pt}
10\% & 90\% & 
48.88	    &	63.03	    &	33.13	\\ \hline
\rowcolor{LightCyan} 50\% & 50\% &
50.75	    &	64.57	    &	32.25	\\ \hline
90\% & 10\% &  
49.19	    &	61.82	    &	34.50	\\ \hline
\specialrule{1.5pt}{0pt}{0pt}
\end{tabular}
\label{tab:ablation_ctr_sampling}
}
\vspace{-15pt}
\end{table}

\subsection{Ablation Study}

We first perform \textbf{an analysis of CAVP components} on AVSBench-Semantics~\cite{zhou2023audio} and VPO(MSMI) in Tab.~\ref{tab:ablation_components}. 
Starting from a baseline consisting of TPAVI-like cross-attention (CA) fusion layer (1st row), we replace this TPAVI CA layer by our CA layer, represented by $f_{CA}(.)$ and observe an average mIoU improvement of around +4.5\% on both datasets. 
Subsequently, by integrating the model with the supervised contrastive learning loss defined in~\cite{khosla2020supervised}, we achieve an additional mIoU improvement of around +1\%, as shown in the 3rd row.
The final row displays CAVP method with all components, including the selection of informative samples, which reaches a further average mIoU improvement of +3.4\%.

\noindent \textbf{Cross-attention.} 
From Sec.~\ref{sec:preliminaries}, recall that $f_{CA}(.)$ can have different activation functions, and we opted for the sigmoid activation to enable the highlighting of multiple regions of varying sizes in the attention map.
We now study the use of different activation functions, namely: 1) softmax~\cite{vaswani2017attention}, 2) channel-wise attention adopted in AVSegFormer~\cite{gao2023avsegformer}, 3) min-max normalisation to produce attention map, and the 4) sigmoid function. 
Results in Fig.~\ref{fig:ablation_ca} show that the setting with the ``sigmoid'' improves the second-best method, ``min-max'', by $+2.38\%$ mIoU.

\noindent \textbf{Sampling Analysis.}
Naturally, the amount of negative samples is overwhelmingly larger than the number of positive samples in the anchor sets of Eq.~\eqref{eq:anchor_sets}, so we need to balance these two sets to enable more effective training. Please refer to the Supplementary Material for details on the balancing process. 
We conduct an ablation study of the model's performance under three different settings containing different proportions of positive and negative samples, as shown in Tab.~\ref{tab:ablation_ctr_sampling}. 
The results reveal that adopting a balanced positive and negative sampling strategy benefits the model's performance, with improvements in mIoU of $1.87\%$ and $1.56\%$ compared to scenarios where either $90\%$ of the samples are negative or $90\%$ are positive.

\begin{figure}[t]
    \centering
     \includegraphics[width=\linewidth]{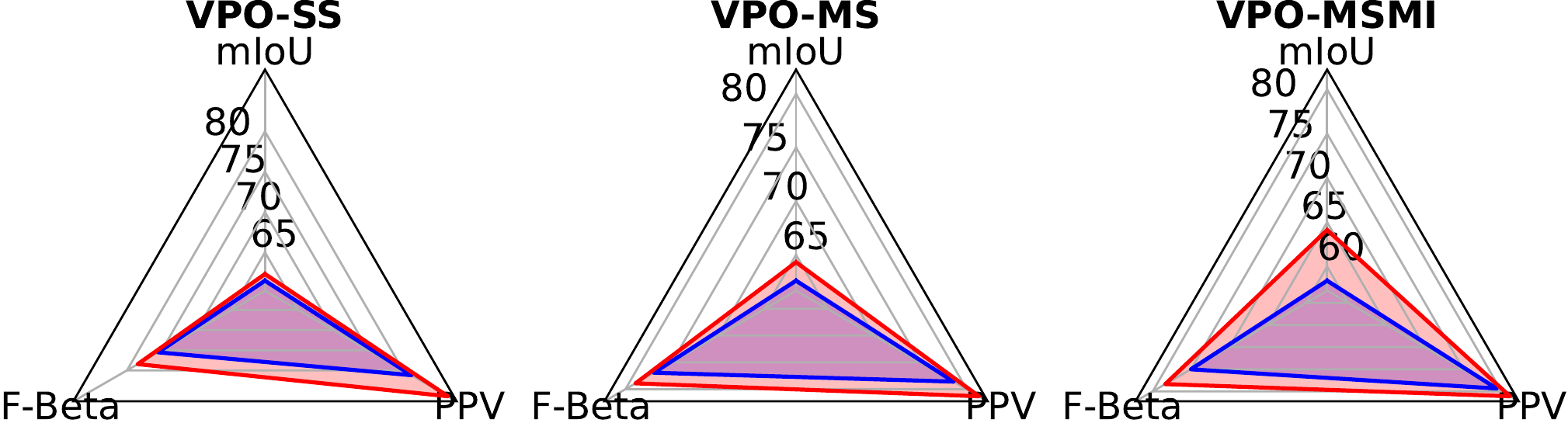}
    \vspace{-10pt}
    \caption{\textcolor{blue}{Mono (blue)} and \textcolor{red}{Stereo (red)} audio VPO performance.}
    \vspace{-15pt}
    \label{fig:ablation_mono_stereo}
\end{figure}

\noindent \textbf{Mono and Stereo Sound.}
To emphasize the significance of stereo sound in the AVS task, we conducted an ablation study on the VPO dataset, toggling the stereo sound on and off, with results shown in Fig.~\ref{fig:ablation_mono_stereo}. Notice that for all VPO subsets, the stereo sound can improve the performance on all evaluation measures\footnote{Note that PPV = 1-FDR in the graph.}.
This improvement is particularly noticeable in the MSMI split as it significantly enhances the model's performance across all three measures. For the numerical results, please refer to the \textit{Supplementary Material}.

\section{Discussion and Conclusion}
In this work, we have presented new cost-effective VPO benchmarks and the innovative CAVP method for Audio-Visual Segmentation (AVS). 
Our proposed VPO benchmarks are both scalable, cost-effective and challenging, while our data collection and annotation protocols provide a substantial reduction of the ``common-sense'' bias found AVS tasks, where certain objects are always the source of sound in some scenarios. 
We also introduce a new supervised audio-visual contrastive learning that utilises arbitrary audio-visual pairs to mine informative contrastive pairs that can better constrain the learning of audio-visual features. Overall, our dataset and method can provide a valuable resource for future research in AVS tasks.\\ 
\noindent\textbf{Limitations and future work.} We recognize that our VPO dataset lacks temporal information and may exhibit a class imbalance issue similar to that observed in AVSBench-Semantics~\cite{zhou2023audio}.
This imbalance results from our strict filtering of trivial cases with training images containing a single dominating-sounding object.
Furthermore, our current approach does not support simulating spatial audio based on objects' visual depth or modelling arrival time differences in the VPO. 
We intend to tackle these issues in our future work to enhance AVS models' robustness and applicability.


\appendix

\begin{figure*}[t]
    \centering
    \begin{subfigure}[b]{.33\linewidth}
         \centering    
            \includegraphics[width=1.\linewidth]{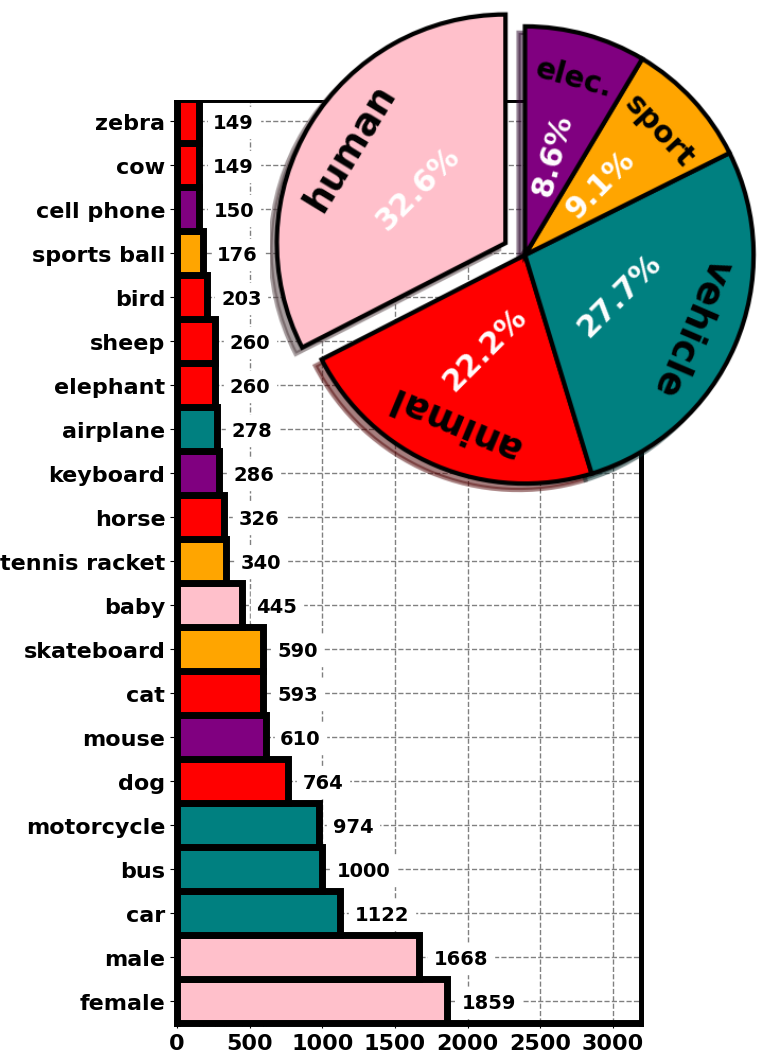}
        \caption{VPO-SS}
        \label{fig:vpo-ss-stat}
         \vspace{-10pt}
    \end{subfigure}
    \begin{subfigure}[b]{.33\linewidth}
        \centering    
            \includegraphics[width=1.\linewidth]{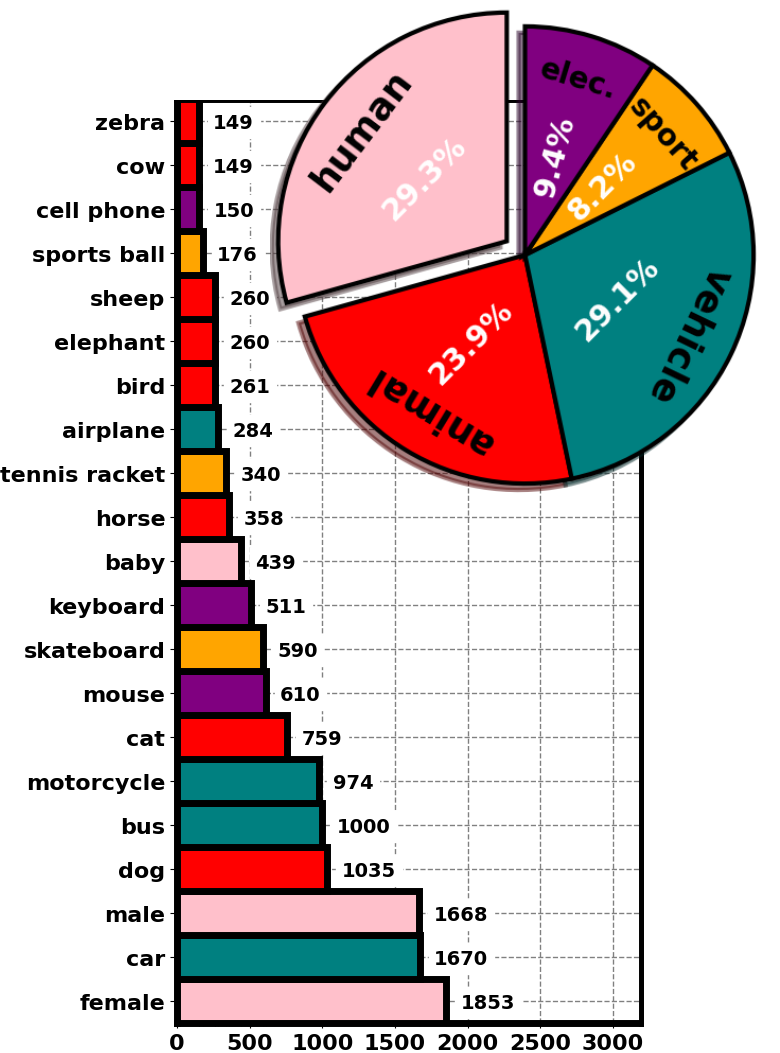}
        \caption{VPO-MS}
        \label{fig:vpo-ms-stat}
         \vspace{-10pt}
    \end{subfigure}
    \begin{subfigure}[b]{.33\linewidth}
        \centering    
            \includegraphics[width=1.\linewidth]{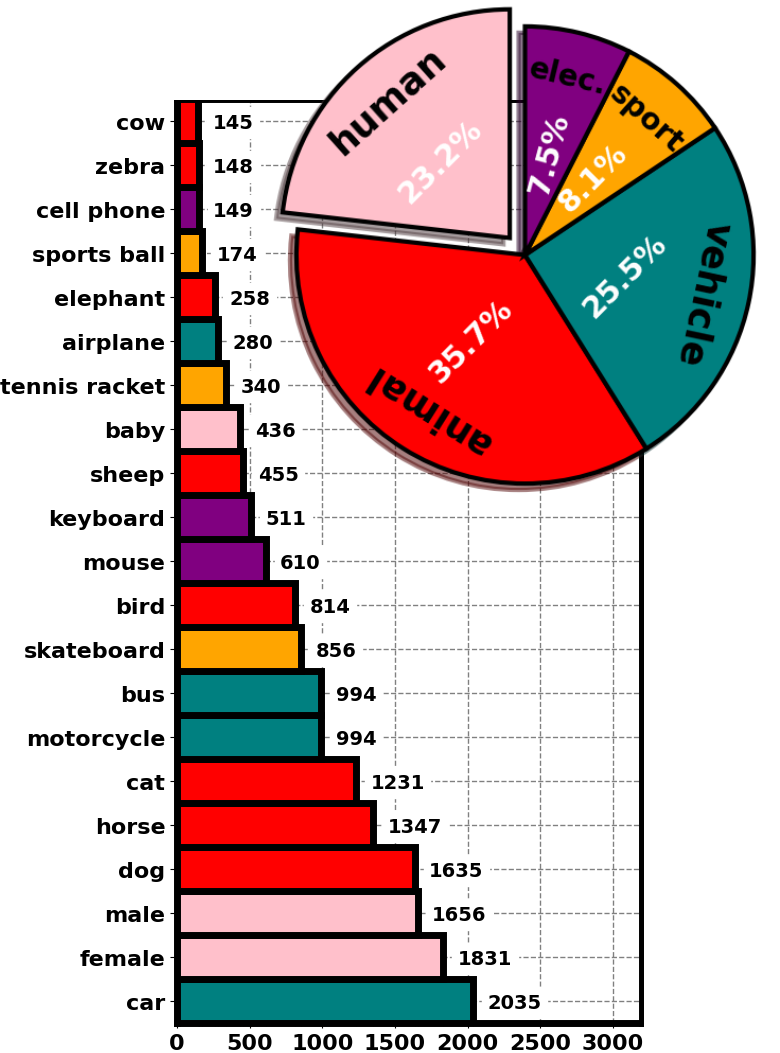}
        \caption{VPO-MSMI}
        \label{fig:vpo-msmi-stat}
         \vspace{-10pt}
    \end{subfigure}
    \caption{Visual class distribution in our proposed VPO-SS, VPO-MS and VPO-MSMI benchmark datasets. 
    }
    \label{fig:data_statistics}
    \vspace{-10pt}
\end{figure*}

\section{Further Details for VPO Dataset}
\subsection{Dataset Statistics}
We show the distribution of visual classes in VPO-SS, VPO-MS and VPO-MSMI in Figure~\ref{fig:data_statistics}. Similar to the AVSBench-Semantics~\cite{zhou2023audio}, we also observe a data imbalance issue within our VPO dataset. We follow~\cite{tang2020long} to report an imbalance ratio ($\frac{N_{\text{max}}}{N_{\text{min}}}$) of 12.48\% (female \& zebra), 12.43\% (female \& zebra) and 12.62\% (female \& cow) on the three VPO subsets, and 59.57\% (man \& axe \& missile-rocket) on AVSBench-Semantics~\cite{zhou2023audio}. 
These class imbalance issues can affect the model performance during testing, which will be discussed in Sec.~\ref{sec:additional_results}.
For the demonstration of training examples, please refer to the ``\textbf{video\_demo.mp4}'' file within the supplementary materials.

\begin{figure*}
    \centering
    \includegraphics[width=\linewidth]{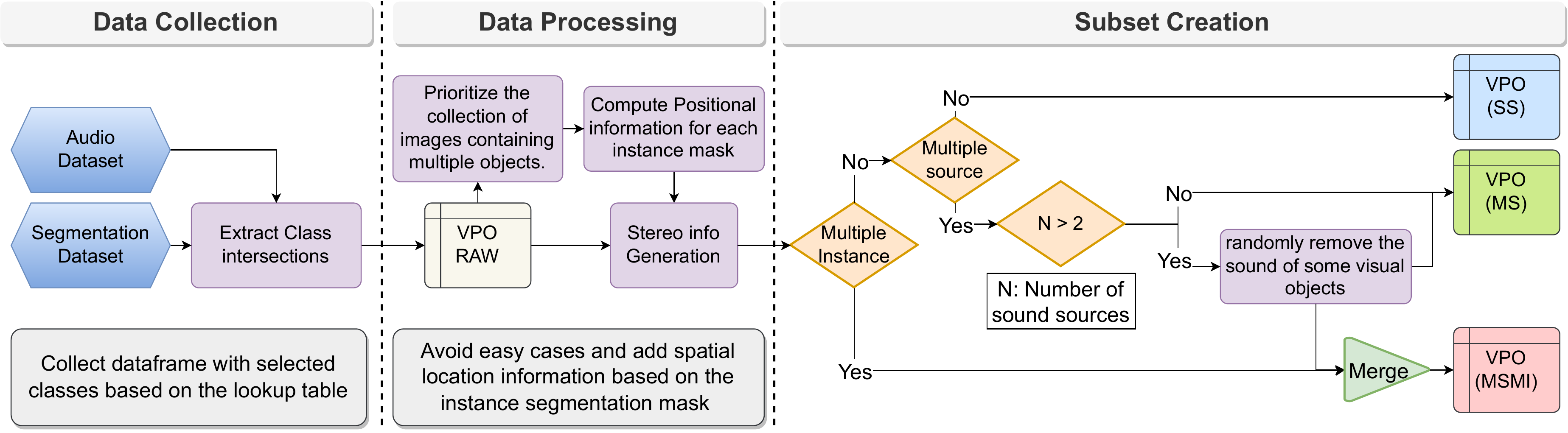}
    \vspace{-10pt}
    \caption{Illustration of VPO collection and production process.}
    \vspace{-10pt}
    \label{fig:vpo-creation}
\end{figure*}

\subsection{Creation Procedure}
We show a graphical illustration of our Visual Post-production (VPO) benchmark in Fig.~\ref{fig:vpo-creation}. We divide the entire dataset generation process into three major steps:
\begin{itemize}
    \item \textbf{Data collection}: We gather datasets from off-the-shelf segmentation datasets (e.g., COCO~\cite{lin2014microsoft}) and audio datasets (e.g., VGGSound~\cite{chen2020vggsound}), focusing on the overlapping classes listed in Tab.~\ref{tab:vpo_lookup_table}. We randomly match audio and video files to form new samples based on their semantic labels.
    \item \textbf{Data processing}: We prioritise the collection of images with multiple objects and incorporate spatial location information based on each selected instance mask.
    \item \textbf{Subset creation}: We organize subsets according to their keywords (e.g., single-source, multi-sources, multi-instances) and further partition each subset into training and testing sets.
\end{itemize}

\begin{table}[t]
    \centering
    \caption{Detailed lookup table correlating audio tags with visual labels in the VPO creation process.}
    \vspace{-8pt}
    \begin{tabular}{c!{\vrule width 1.5pt}c}
    \specialrule{1.5pt}{0pt}{0pt}
        Visual-label  & Audio-label \\ 
    \specialrule{1.5pt}{0pt}{0pt}
        bird  & mynah bird singing \\ \hline
        keyboard  & typing on computer keyboard \\ \hline
        bus & driving buses \\ \hline
        cat & \makecell{cat purring / cat purring  \\ cat meowing / cat caterwaulin} \\ \hline
        dog & \makecell{dog growling / dog bow-wow \\ dog whimpering / dog howling \\ dog barking / dog baying} \\ \hline
        horse & horse neighing / horse clip-clop \\ \hline
        car & \makecell{car passing by  / car engine idling \\ car engine starting \\ race car, auto racing \\  car engine knocking} \\ \hline
        sports ball & shot football \\ \hline
        airplane & airplane / airplane flyby \\ \hline
        sheep & sheep bleating \\ \hline
        cow & cow lowing \\ \hline
        motorcycle & driving motorcycle \\  \hline
        mouse & mouse clicking \\ \hline
        cell phone & cell phone buzzing \\ \hline
        elephant & elephant trumpeting \\ \hline
        zebra & zebra braying \\ \hline
        tennis racket & playing tennis \\ \hline
        skateboard & skateboarding \\ \hline
        male & \makecell{male speech, man speaking \\ male singing} \\ \hline
        female & female speech, woman speaking \\ \hline
        baby & \makecell{baby babbling / baby crying \\ baby laughter} \\ \hline
    \specialrule{1.5pt}{0pt}{0pt}
    \end{tabular}
    \vspace{-15pt}
    \label{tab:vpo_lookup_table}
\end{table}

\subsection{Strengths and Weaknesses of VPO}
As we discussed in the main paper, our VPO dataset enjoys the following \textbf{strengths}:
\begin{itemize}
    \item \emph{\underline{Cost-effectiveness}}. The manually labelled AVS dataset needs annotators to watch and listen to an entire video so they can provide labels. The VPO production process can significantly reduce such annotation costs.
    \item \emph{\underline{Data scalability}}. We can easily increase the amount of data by leveraging existing visual (ADE20K~\cite{zhou2017scene}, Pascal VOC~\cite{everingham2010pascal}, etc.) and audio (AudioSet~\cite{gemmeke2017audio}, ESC-50~\cite{piczak2015esc}, etc.) datasets.
    \item \emph{\underline{More diverse scenarios.}} In our proposed VPO datasets, every object within the scene will have a chance to be the sounding object, which is crucial to mitigate the ``commonsense'' bias that is observed in AVSBench-Semantics~\cite{zhou2023audio} and addressing the assessment of spurious correlation~\cite{liu2019clevr,arjovsky2019invariant,cirik2018visual}.

    \item \emph{\underline{Isolation of motion information}.} The disentanglement of motion and sound in our VPO benchmark prevents the model from solely relying on motion information to make predictions, encouraging the learning and evaluation of cross-modal alignment~\cite{senocak2023sound}.
    \item \emph{\underline{Introduction of stereo audio}.} The use of stereo audio encourages the study of spatial prompts.
\end{itemize}
We also identified the following \textbf{weaknesses} of VPO:
\begin{itemize}
    \item \emph{\underline{Data imbalance issue}.} Such imbalance can affect the segmentation accuracy, particularly for the tail classes.
    \item \emph{\underline{Lack of temporal image data}.} Since we match still images with audio, we cannot use image motion information.
    \item \emph{\underline{Comprehensive simulations of spatial audio}.} Our VPO does not include the modelling of arrival time differences and microphone distance.
\end{itemize}

\begin{figure*}[!ht]
    \centering
    \includegraphics[width=1\linewidth]{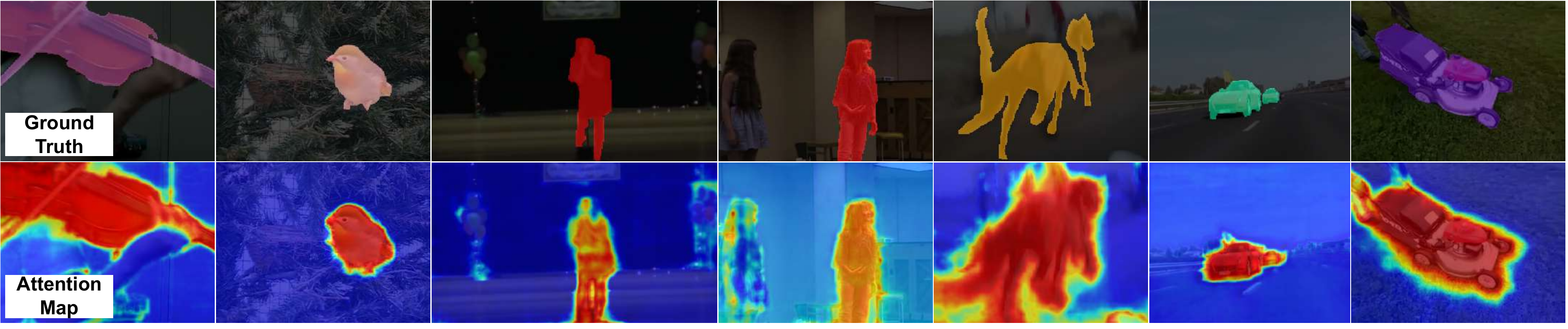}
    \vspace{-10pt}
    \caption{Qualitative results for cross-attention heatmap on AVSBench-Semantics~\cite{zhou2023audio}.}
    \vspace{-10pt}
    \label{fig:visual_attn}
\end{figure*}

\begin{figure*}[!ht]
\centering
    \centering
    \begin{subfigure}[b]{.245\linewidth}
         \centering    
            \includegraphics[width=1.\linewidth]{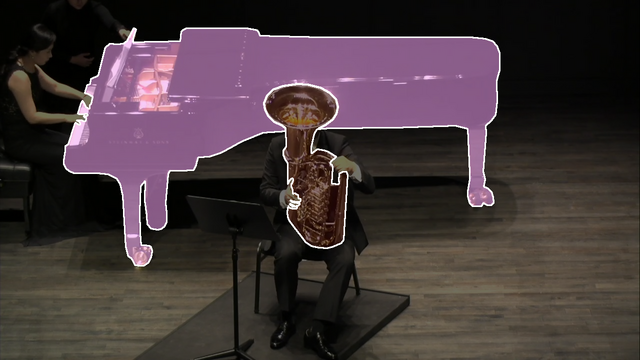}
        \caption{Ground Truth}
        \label{fig:attn_gt}
         \vspace{-10pt}
    \end{subfigure}
    \begin{subfigure}[b]{.245\linewidth}
        \centering    
            \includegraphics[width=1.\linewidth]{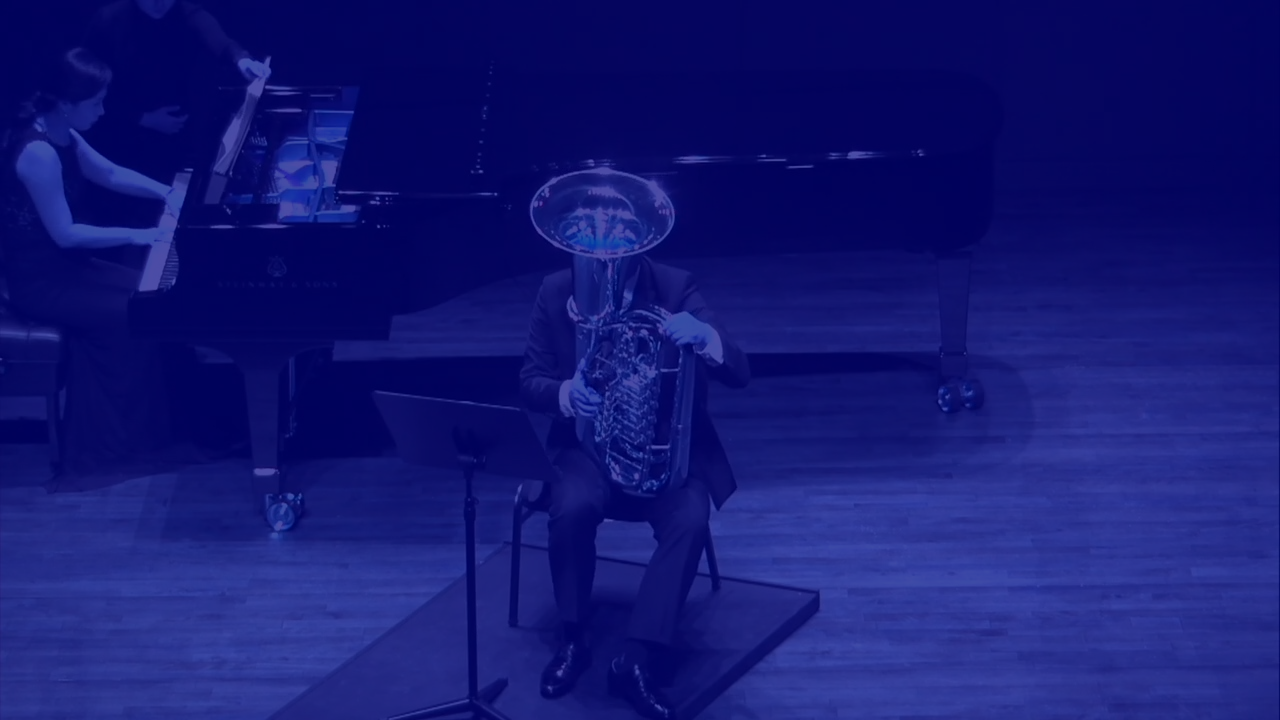}
        \caption{Softmax}
        \label{fig:attn_softmax}
         \vspace{-10pt}
    \end{subfigure}
    \begin{subfigure}[b]{.245\linewidth}
        \centering    
            \includegraphics[width=1.\linewidth]{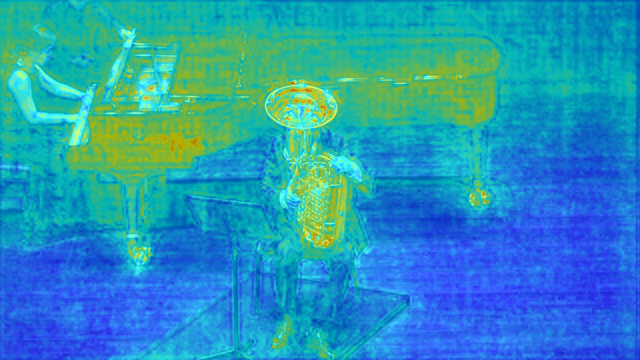}
        \caption{Min-max}
        \label{fig:attn_minmax}
         \vspace{-10pt}
    \end{subfigure}
    \begin{subfigure}[b]{.245\linewidth}
        \centering    
            \includegraphics[width=1.\linewidth]{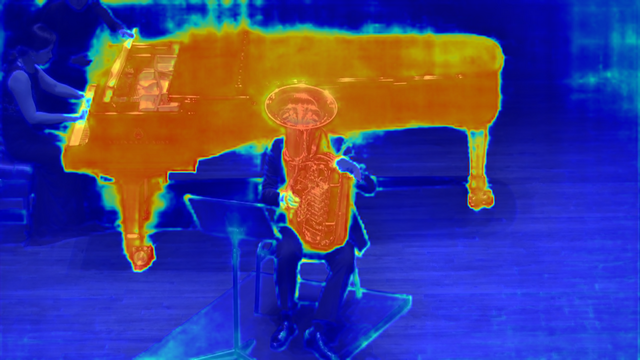}
        \caption{Sigmoid}
        \label{fig:attn_sigmoid}
         \vspace{-10pt}
    \end{subfigure}
    \caption{Visualisation of attention map process by Softmax and Sigmoid activation.}
    \vspace{-10pt}
    \label{fig:attn-comparison}
\end{figure*}

\begin{figure*}[!ht]
    \centering
    \begin{subfigure}[b]{.495\linewidth}
         \centering    
            \includegraphics[width=1.\linewidth]{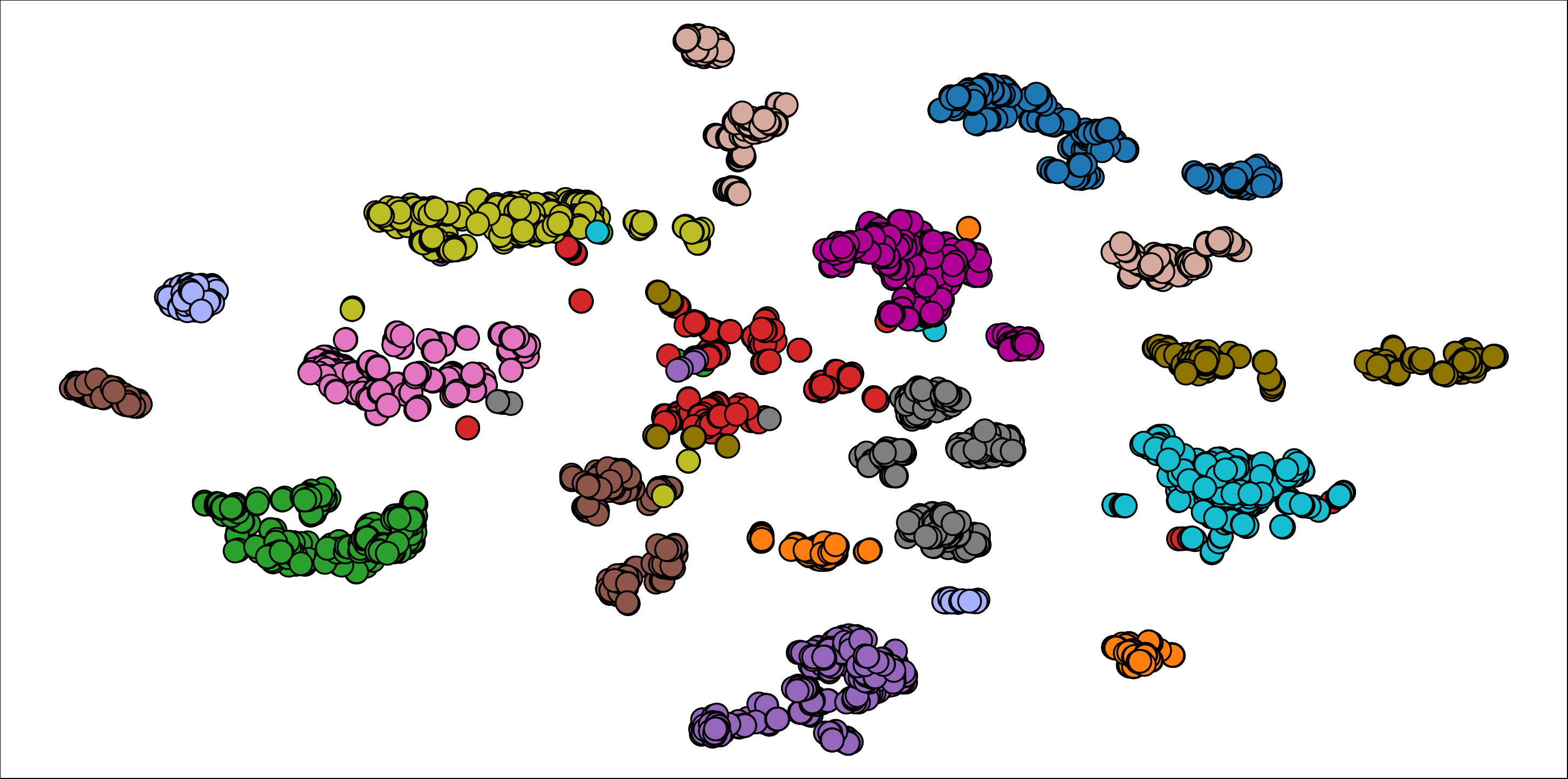}
        \caption{Without CAVP}
        \label{fig:tsne_wo_cavp}
        \vspace{-10pt}
    \end{subfigure}
    \begin{subfigure}[b]{.495\linewidth}
        \centering    
            \includegraphics[width=1.\linewidth]{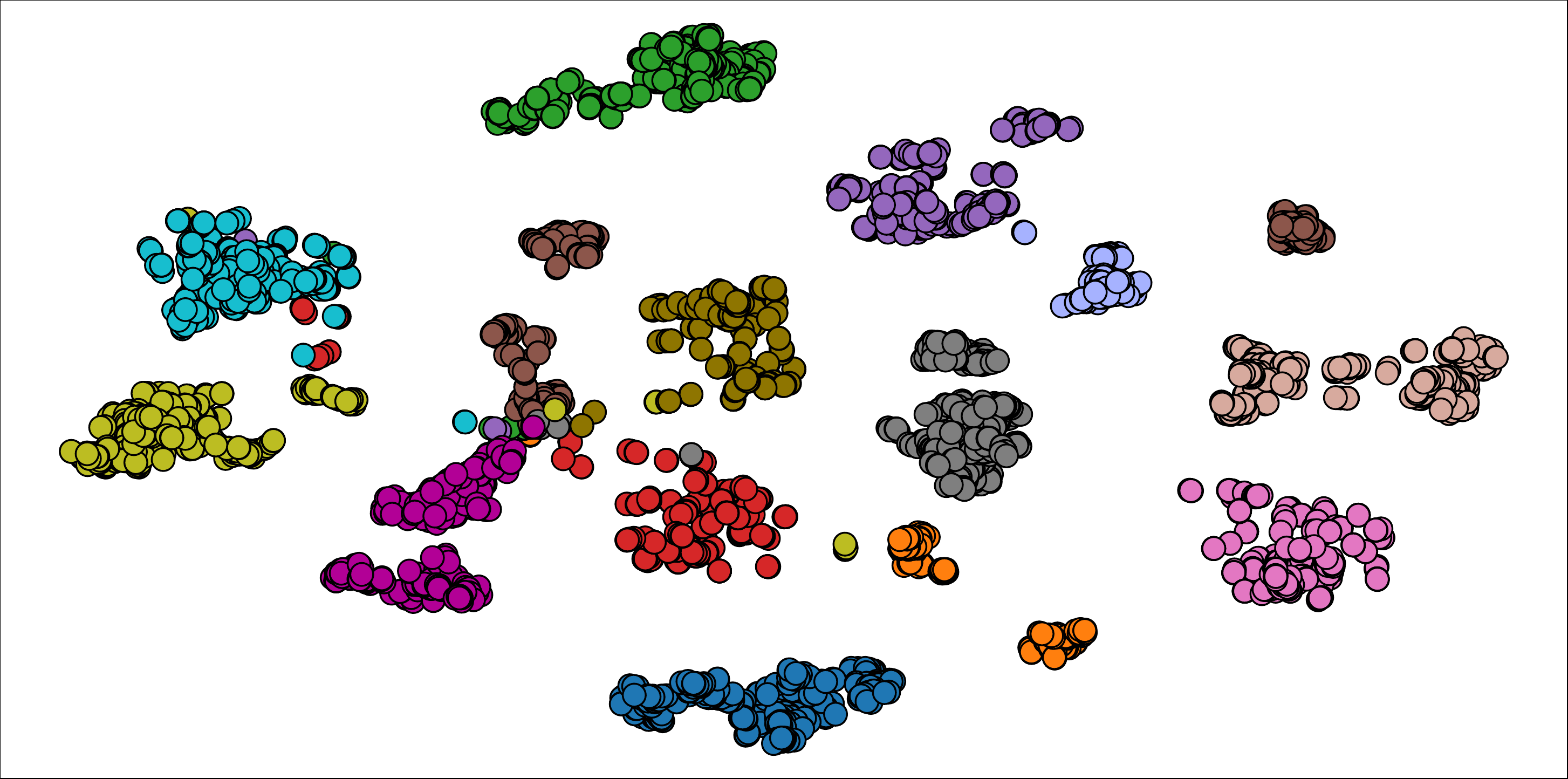}
        \caption{With CAVP}
        \label{fig:tsne_w_cavp}
        \vspace{-10pt}
    \end{subfigure}
    \caption{T-SNE visualisation of features after cross-attention fusion layer, trained with or without our CAVP.}
    \vspace{-10pt}
    \label{fig:tsne_avss}
\end{figure*}

\section{Experiments}

\subsection{Implementation Details}
\noindent \textbf{Training \& Inference:} During training, we apply data augmentation for image inputs with colour jitter, horizontal flipping and random scaling between 0.5 to 2.0. We randomly crop images to $512\times512$ pixels.
For the audio data, we extract the log Mel-Spectrogram with
64 mel filter banks over $1$s~\cite{zhou2022audio} or $3$s~\cite{mo2022closer} of the waveform at 16 kHz on AVSBench and VPO. We set temperature $\tau$ as 0.1.
We use the SGD optimizer with a momentum of 0.9, weight decay of 0.0005, and a polynomial learning-rate decay 
$(1-\frac{\text{iter}}{\text{total\_iter}})^\text{power}$ with $\text{power}=0.9$. The initial learning rate is set to $0.001$, the mini-batch size is 16 and training lasts for 80 epochs. During inference, we use the original resolution resizing and cropping with a mini-batch size of 1. 
We adopt two image backbones (ResNet~\cite{he2016deep}, PVT-V2-B5~\cite{wang2022pvt}) and DeepLabV3+~\cite{chen2018encoder} for the segmentation network.
For the audio backbones, 
we use VGGish~\cite{hershey2017cnn} (following~\cite{zhou2022audio}) and ResNet-18~\cite{he2016deep} (following~\cite{chen2021localizing, mo2022closer}) for AVSBench and VPO, respectively.
The overwhelming amount of negative samples is mitigated by maintaining a memory bank~\cite{he2020momentum} to store raw waveform data for each class. During training, we proportionally transfer negative pairs from the negative set to the positive set by pairing positive audio with respect to the image label.

\begin{table*}[t]
    \centering
    \caption{Quantitative (mIoU, $F_{\beta}$, FDR) audio-visual segmentation results (in \%) on VPO dataset with ResNet50~\cite{he2016deep} backbone and mono audio. Best results are in \textbf{bold}, and second best are\underline{underlined}. Improvements against the second best are in the last row.}
    \label{tab:vpo_mono_resnet50}
    \vspace{-5pt}
    \rowcolors{5}{LightCyan}{LightCyan}
    \resizebox{\linewidth}{!}{
    \begin{tabular}{!{\vrule width 1.5pt}c!{\vrule width 1.5pt}c!{\vrule width 1.5pt}ccc|ccc|ccc!{\vrule width 1.5pt}} 
    \hline
\specialrule{1.5pt}{0pt}{0pt}
\multirow{2}{*}{D-ResNet50~\cite{he2016deep}} & \multirow{2}{*}{Method} & \multicolumn{3}{c|}{VPO (SS)} & \multicolumn{3}{c|}{VPO (MS)} & \multicolumn{3}{c!{\vrule width 1.5pt}}{VPO (MSMI)} \\
\cline{3-11}
\rule{0pt}{10pt}
~ & ~ &  mIoU $\uparrow$ & $F_{\beta}$ $\uparrow$ & FDR 
$\downarrow$ & mIoU $\uparrow$ & $F_{\beta}$ $\uparrow$ & FDR $\downarrow$ & mIoU $\uparrow$ & $F_{\beta}$ $\uparrow$ & FDR $\downarrow$ \\ 
\specialrule{1.5pt}{0pt}{0pt}
\global\let\CT@@do@color\relax 
\multirow{1}{*}{Transformer} & 
AVSegFormer~\cite{gao2023avsegformer} & 
52.96 & 67.89 & 25.50 & 
56.46 & 71.89 & 24.66 & 
50.96 & 64.96 & 32.72 \\ \hline
\multirow{2}{*}{Per-pixel Classification} & 
TPAVI~\cite{zhou2022audio} & 
51.84 & 68.77 & 23.64 & 
44.08 & 58.14 & 30.82 & 
50.37 & 66.80 & 29.82 \\ 
& \global\let\CT@@do@color\oriCT@@do@color\textbf{Ours} & 
\textbf{61.48} & \textbf{75.53} & \textbf{18.79} & 
\textbf{61.85} & \textbf{74.60} & \textbf{20.24} & 
\textbf{57.22} & \textbf{72.26} & \textbf{24.04} \\ \hline
Improvements & \textbf{Ours} & 
\improve{+8.52} & \improve{+6.76} & \drop{-4.85} & 
\improve{+5.39} & \improve{+2.71} & \drop{-4.42} & 
\improve{+6.26} & \improve{+5.46} & \drop{-5.78} \\
\specialrule{1.5pt}{0pt}{0pt}
    \end{tabular}
}
\vspace{-10pt}
\end{table*}

\section{Attention Map Visualisation}

To demonstrate the effectiveness of our cross-attention module, we visualize the audio-visual attention heatmap. We employ models pre-trained on AVSBench-Semantics~\cite{zhou2023audio}, utilizing a full-resolution set-up and equipped with a ResNet50~\cite{he2016deep} backbone. 
As shown in Fig.~\ref{fig:visual_attn}, the heatmaps illustrate that our module can effectively retrieve the foreground object by leveraging the interaction between audio and visual embeddings.
Additionally, we show a visual comparison amount of the application of $\mathrm{softmax}(.)$, $\mathrm{Minmax}(.)$ and $\mathrm{sigmoid}(.)$ activation functions. 
As depicted in Fig.~\ref{fig:attn-comparison}, using Softmax under spatial dimension may lead to a diminished attention map (Fig.~\ref{fig:attn_softmax}), while incorporating Minmax activation over the dot product of the audio-visual feature could yield a noisy and limited discernment of relevant audio-visual correspondences. (Fig.~\ref{fig:attn_minmax}). Our method in Fig.~\ref{fig:attn_sigmoid}, utilizing $\mathrm{sigmoid}(.)$, demonstrates a better efficacy compared to these two methods in terms of cross-modal feature activation.

\subsection{T-SNE Visualisation}

To present the qualitative results of our CAVP method in the latent space, we extract the features computed before the classification layers, and generate T-SNE plots in Fig.~\ref{fig:tsne_avss}. We employ models pre-trained on AVSBench-Semantics~\cite{zhou2023audio}, utilizing a full-resolution set-up and equipped with a ResNet50~\cite{he2016deep} backbone. We consider both scenarios: with the proposed CAVP method (Fig.~\ref{fig:tsne_w_cavp}) and without it (Fig.~\ref{fig:tsne_wo_cavp}). The results demonstrate that our method can enhance intra-class compactness while preserving intra-class separability.




\subsection{Additional Results} 
\label{sec:additional_results}
We present supplementary results for the paper, showcasing the performance of AVSegformer~\cite{gao2023avsegformer}, TPAVI~\cite{zhou2022audio}, and our model on \textbf{VPO with mono audio} with ResNet50~\cite{he2016deep} backbone, as depicted in Tab.~\ref{tab:vpo_mono_resnet50}. Our method outperforms the baseline methods on all experimental settings by a minimum of 5.39\%, 2.71\% and 4.42\% for mIoU, $F_{\beta}$ score and false detection rate, respectively.
We also provide \textbf{class-wise results} on AVSBench-Semantics~\cite{zhou2023audio} in Tables~\ref{tab:avss_class_miou},\ref{tab:avss_class_fscore},\ref{tab:avss_class_fdr}. We observe that the tail classes, such as clippers, axe, missile-rocket and utv, show significantly worse results than the remaining classes, illustrating the importance of addressing the imbalance issue in the AVS task.

\begin{figure}[!ht]
    \centering
    \includegraphics[width=\linewidth]{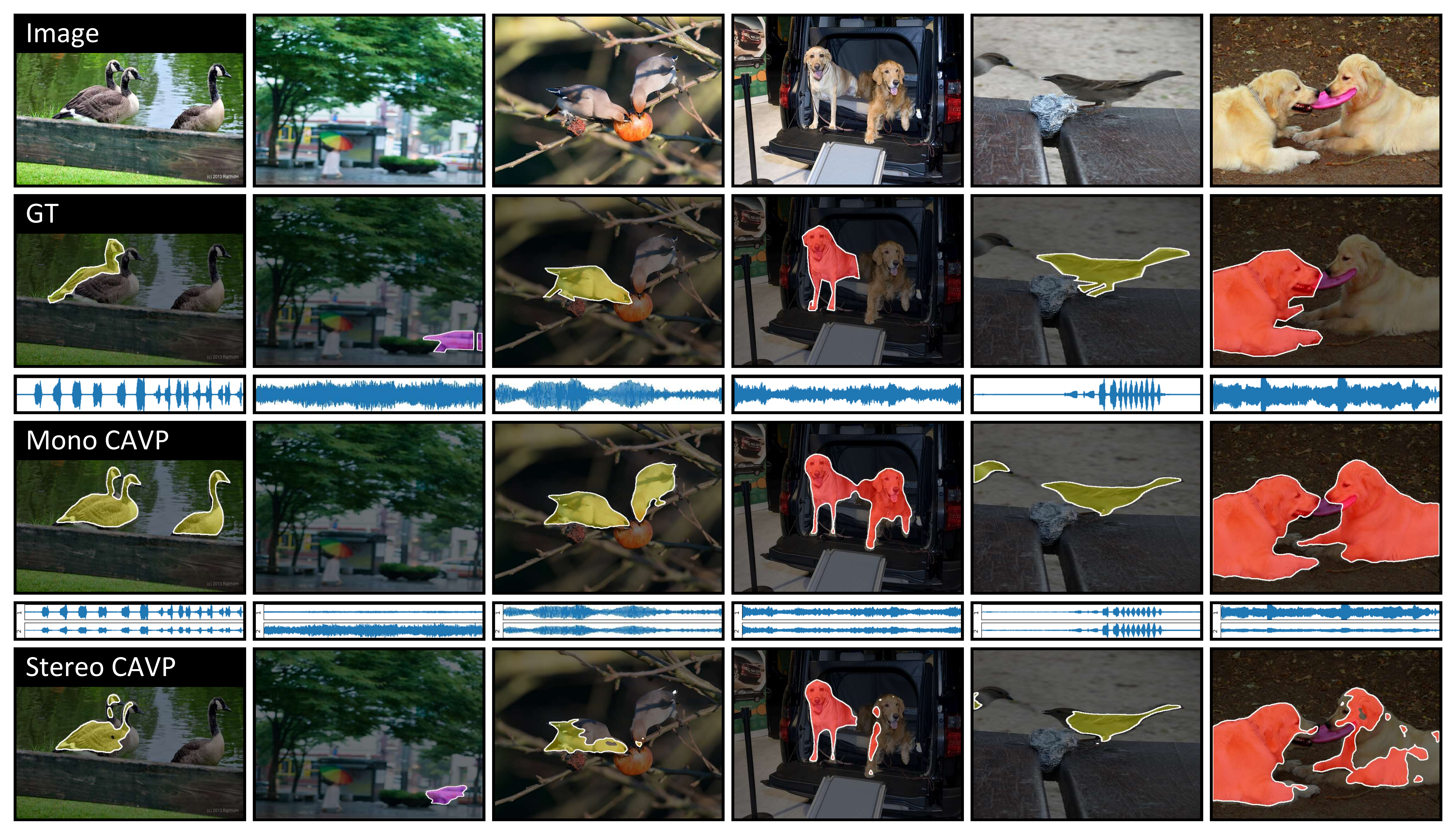}
    \vspace{-8pt}
    \caption{Visual comparison between CAVP models trained with mono audio and stereo audio.}
    \vspace{-15pt}
    \label{fig:mono_vs_stereo}
\end{figure}

To showcase the model's performance with both mono and stereo audio inputs, we opted for two CAVP models utilizing the ResNet50~\cite{he2016deep} backbone. Trained on VPO-MSMI, these models were chosen to assess prediction outcomes in multi-instance scenarios, illustrated in Fig.~\ref{fig:mono_vs_stereo}. 
Our observations reveal that, when aided by stereo audio, the model effectively diminishes its focus on the incorrect spatial direction. 
However, challenges persist in handling images with multiple instances, exemplified in the last column (depicting two dogs) of Fig.~\ref{fig:mono_vs_stereo}. This underscores a significant challenge in the audio-visual segmentation (AVS) task.

\begin{figure}[h]
    \centering
    \includegraphics[width=1\linewidth]{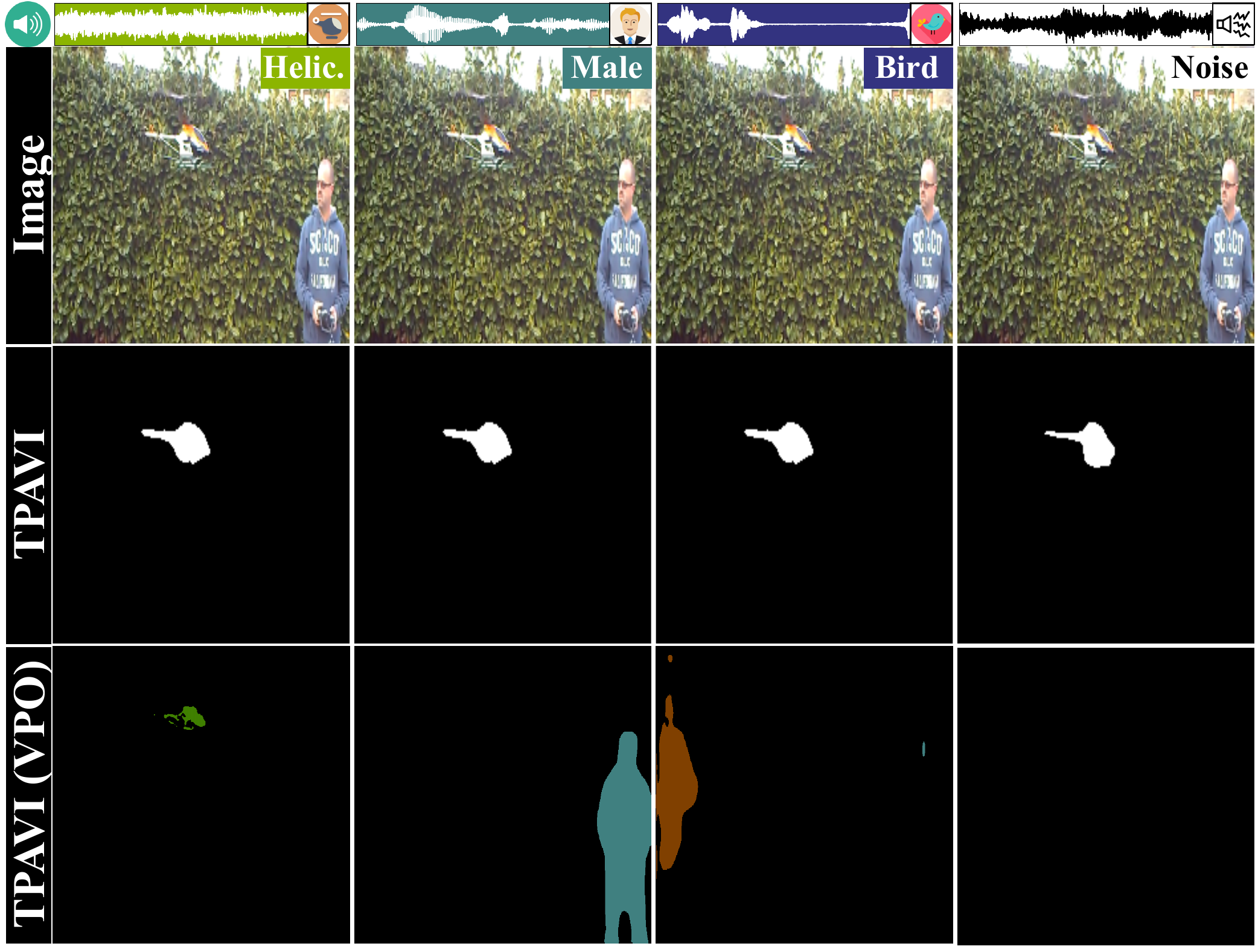}
    \vspace{-8pt}
    \caption{
    Visual comparison between TPAVI models trained on AVSBench~\cite{zhou2022audio} (2nd row) and VPO (3rd row), using an identical set of synthetic test samples. The columns display the original audio (1st column) and alternative sound types, including a person speaking (2nd column), bird chirping (3rd column), or background noise (4th column).
    }
    \vspace{-15pt}
    \label{fig:avs-motivation_rebuttal}
\end{figure}
To further demonstrate the effectiveness of our VPO dataset, we used the TPAVI~\cite{zhou2022audio} trained on VPO and applied it to test the synthetic examples, shown in Fig.~\ref{fig:avs-motivation_rebuttal}. Despite hallucinations persisting in the 3rd column (class ``Bird''), there is a noticeable improvement, with the correct segment of ``Helicopter'', ``Male'', and ``Noise'', compared to the results in the 2nd row.

\begin{figure*}[t]
    \centering
    \includegraphics[width=\linewidth]{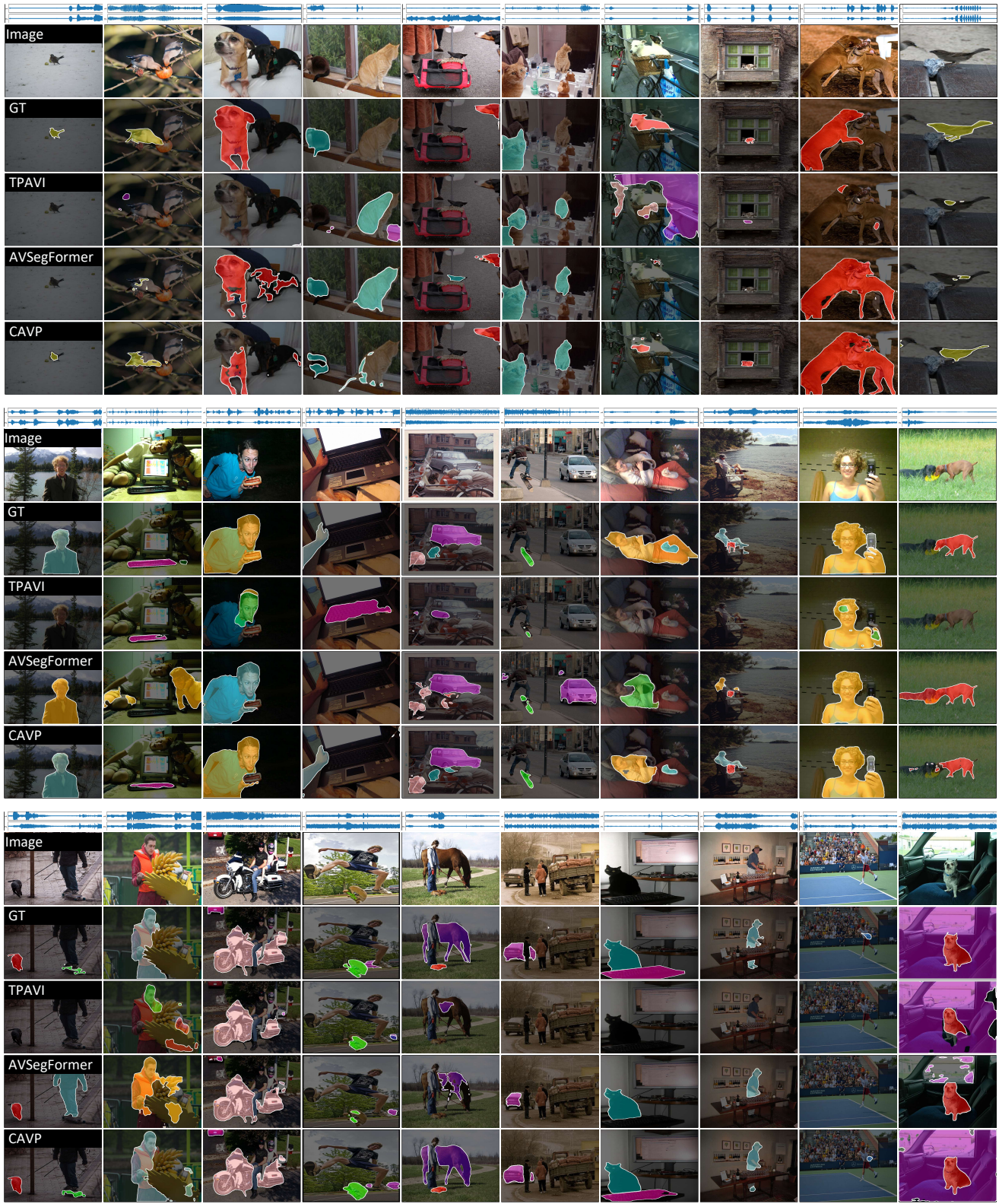}
    \caption{Qualitative audio-visual segmentation results on VPO by TPAVI~\cite{zhou2022audio}, AVSegFormer~\cite{gao2023avsegformer}, and our CAVP. The prediction results can be compared with the ground truth (GT) of the first row of each sample.}
    \label{fig:vpo_visual_all}
\end{figure*}

\begin{figure*}[t]
    \centering
    \includegraphics[width=\linewidth]{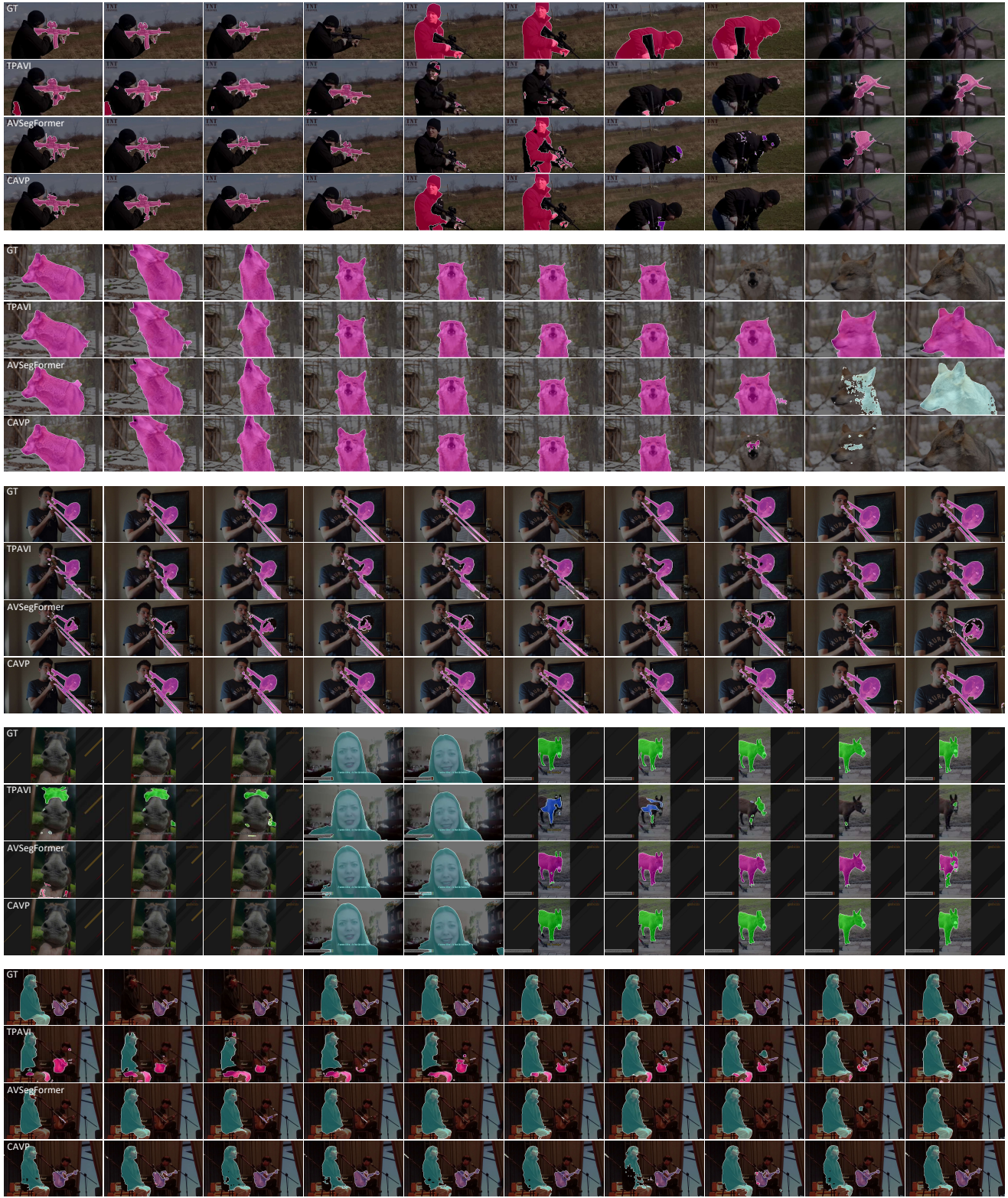}
    \caption{Qualitative audio-visual segmentation results on AVSBench-Semantics~\cite{zhou2023audio} by TPAVI~\cite{zhou2022audio}, AVSegFormer~\cite{gao2023avsegformer}, and our CAVP. The prediction results can be compared with the ground truth (GT) of the first row of each video.
    }
    \label{fig:avss_visual_all}
\end{figure*}

Finally, we show a qualitative comparison visualization among TPAVI~\cite{zhou2022audio}, AVSegFormer~\cite{gao2023avsegformer}, and our CAVP on VPO in Fig.~\ref{fig:vpo_visual_all} and on AVSBench-Semantics in Fig.~\ref{fig:avss_visual_all}. We demonstrate that our method consistently provides a more effective approximation of the true segmentation of objects in the scene compared to alternative methods.
For the demonstration of full video examples on AVSBench-Semantics, please refer to the ``\textbf{video\_demo.mp4}'' file within the attached supplementary materials.

\begin{table*}[!ht]
    \centering
    \caption{Class-level (mIoU) audio-visual segmentation results (in \%) on AVSBench-Semantic dataset~\cite{zhou2023audio} (original resolution) with ResNet50~\cite{he2016deep} backbone. 
    }
    \vspace{-5pt}
    \label{tab:avss_class_miou}
    \def\arraystretch{1.1}
    \resizebox{\linewidth}{!}{
    \begin{tabular}{!{\vrule width 1.5pt}c!{\vrule width 1.5pt}ccccccccccccccc!{\vrule width 1.5pt}} 
\specialrule{1.5pt}{0pt}{0pt}
Class & background & accordion & airplane & axe & baby & bassoon & bell & bird & boat & boy & bus & car & cat & cello & clarinet \\ 
\specialrule{1.5pt}{0pt}{0pt}
AVSegformer & 0.9093 & 0.8610 & 0.9003 & 0.0000 & 0.4178 & 0.1683 & 0.0618 & 0.2251 & 0.7751 & 0.0130 & 0.4240 & 0.2800 & 0.3952 & 0.4553 & 0.0033 \\ 
TPAVI & 0.9111 & 0.7484 & 0.9261 & 0.0000 & 0.4147 & 0.1578 & 0.0795 & 0.3814 & 0.7149 & 0.0251 & 0.7019 & 0.4214 & 0.5209 & 0.4537 & 0.1624 \\ 
\textbf{CAVP} & 0.9168 & 0.9229 & 0.9291 & 0.0000 & 0.4212 & 0.2580 & 0.3376 & 0.2481 & 0.8096 & 0.1194 & 0.3486 & 0.3786 & 0.4968 & 0.5490 & 0.0180 \\ 
\specialrule{1.5pt}{0pt}{0pt}
Class & clipper & clock & dog & donkey & drum & duck & elephant & emergency-car & erhu & flute & frying-food & girl & goose & guitar & ~ \\ 
\specialrule{1.5pt}{0pt}{0pt}
AVSegformer & 0.0000 & 0.5482 & 0.2847 & 0.3876 & 0.0288 & 0.5581 & 0.8649 & 0.5377 & 0.2396 & 0.2287 & 0.5187 & 0.1163 & 0.2171 & 0.7520 & ~ \\ 
TPAVI & 0.0000 & 0.4179 & 0.2329 & 0.1913 & 0.3180 & 0.4620 & 0.7492 & 0.5342 & 0.3697 & 0.3134 & 0.4161 & 0.1737 & 0.0054 & 0.7203 & ~ \\ 
\textbf{CAVP} & 0.0000 & 0.7088 & 0.3259 & 0.4141 & 0.2987 & 0.5154 & 0.7597 & 0.6329 & 0.5898 & 0.3358 & 0.4761 & 0.1691 & 0.6841 & 0.8172 & ~ \\ 
\specialrule{1.5pt}{0pt}{0pt}
Class & gun & guzheng & hair-dryer & handpan & harmonica & harp & helicopter & hen & horse & keyboard & leopard & lion & man & marimba & ~ \\ 
\specialrule{1.5pt}{0pt}{0pt}
AVSegformer & 0.2456 & 0.4480 & 0.4926 & 0.8611 & 0.0000 & 0.6583 & 0.6952 & 0.1423 & 0.1709 & 0.7823 & 0.6721 & 0.7946 & 0.4050 & 0.8194 & ~ \\ 
TPAVI & 0.2355 & 0.4999 & 0.6422 & 0.7801 & 0.0000 & 0.5573 & 0.7185 & 0.2600 & 0.3104 & 0.7592 & 0.6018 & 0.7331 & 0.4229 & 0.8223 & ~ \\ 
\textbf{CAVP} & 0.3059 & 0.5587 & 0.6075 & 0.8885 & 0.0000 & 0.6473 & 0.7762 & 0.7078 & 0.2506 & 0.7800 & 0.6797 & 0.8328 & 0.4300 & 0.8392 & ~ \\ 
\specialrule{1.5pt}{0pt}{0pt}
Class & missile-rocket & motorcycle & mower & parrot & piano & pig & pipa & saw & saxophone & sheep & sitar & sorna & squirrel & tabla & ~ \\ 
\specialrule{1.5pt}{0pt}{0pt}
AVSegformer & 0.0000 & 0.0501 & 0.6662 & 0.1493 & 0.6031 & 0.2891 & 0.5213 & 0.2953 & 0.3404 & 0.0837 & 0.4364 & 0.2090 & 0.4190 & 0.6088 & ~ \\ 
TPAVI & 0.1086 & 0.0106 & 0.6237 & 0.1239 & 0.5406 & 0.2505 & 0.6602 & 0.4154 & 0.4560 & 0.0507 & 0.5225 & 0.6431 & 0.4567 & 0.6963 & ~ \\ 
\textbf{CAVP} & 0.0000 & 0.1063 & 0.6487 & 0.1158 & 0.6178 & 0.6247 & 0.6518 & 0.5625 & 0.5141 & 0.1069 & 0.6552 & 0.4723 & 0.6357 & 0.6383 & ~ \\ 
\specialrule{1.5pt}{0pt}{0pt}
Class & tank & tiger & tractor & train & trombone & truck & trumpet & tuba & ukulele & utv & vacuum-cleaner & violin & wolf & woman & ~ \\ 
\specialrule{1.5pt}{0pt}{0pt}
AVSegformer & 0.3381 & 0.5467 & 0.4995 & 0.8996 & 0.4133 & 0.2434 & 0.3514 & 0.7967 & 0.5930 & 0.0000 & 0.1720 & 0.4008 & 0.7839 & 0.3837 & ~ \\ 
TPAVI & 0.4718 & 0.5849 & 0.4586 & 0.8529 & 0.4425 & 0.1831 & 0.2631 & 0.7744 & 0.5388 & 0.0000 & 0.1558 & 0.5302 & 0.6924 & 0.4285 & ~ \\ 
\textbf{CAVP} & 0.6514 & 0.6226 & 0.4922 & 0.9427 & 0.6415 & 0.1948 & 0.3839 & 0.8249 & 0.7038 & 0.0000 & 0.5680 & 0.6005 & 0.8510 & 0.4157 & ~ \\ 
\specialrule{1.5pt}{0pt}{0pt}
\end{tabular}
}
\vspace{-10pt}
    
\end{table*}
\begin{table*}[!ht]
    \centering
    \caption{Class-level ($F_{\beta}$) audio-visual segmentation results (in \%) on AVSBench-Semantic dataset~\cite{zhou2023audio} (original resolution) with ResNet50~\cite{he2016deep} backbone.
    }
    \vspace{-5pt}
    \label{tab:avss_class_fscore}
    \def\arraystretch{1.1}
    \resizebox{\linewidth}{!}{
    \begin{tabular}{!{\vrule width 1.5pt}c!{\vrule width 1.5pt}ccccccccccccccc!{\vrule width 1.5pt}} 
\specialrule{1.5pt}{0pt}{0pt}
Class & background & accordion & airplane & axe & baby & bassoon & bell & bird & boat & boy & bus & car & cat & cello & clarinet \\ 
\specialrule{1.5pt}{0pt}{0pt}
AVSegformer & 0.9484 & 0.9395 & 0.9488 & 0.0000 & 0.5795 & 0.4054 & 0.1657 & 0.2971 & 0.8914 & 0.0351 & 0.5099 & 0.3539 & 0.5359 & 0.6597 & 0.0129 \\ 
TPAVI & 0.9445 & 0.8090 & 0.9581 & 0.0000 & 0.6223 & 0.4018 & 0.2440 & 0.5241 & 0.8458 & 0.0786 & 0.8474 & 0.5883 & 0.7307 & 0.6560 & 0.3185 \\ 
\textbf{CAVP} & 0.9528 & 0.9579 & 0.9505 & 0.0000 & 0.6393 & 0.5019 & 0.5094 & 0.3887 & 0.8846 & 0.2930 & 0.4237 & 0.4606 & 0.6165 & 0.7143 & 0.0535 \\ 
%
\specialrule{1.5pt}{0pt}{0pt}
Class & clipper & clock & dog & donkey & drum & duck & elephant & emergency-car & erhu & flute & frying-food & girl & goose & guitar & ~ \\ 
\specialrule{1.5pt}{0pt}{0pt}
AVSegformer & 0.0000 & 0.7194 & 0.4178 & 0.6265 & 0.0956 & 0.7510 & 0.9100 & 0.6769 & 0.5244 & 0.3931 & 0.7594 & 0.2153 & 0.4183 & 0.8626 & ~ \\ 
TPAVI & 0.0000 & 0.6666 & 0.4233 & 0.4442 & 0.6056 & 0.6664 & 0.8798 & 0.7391 & 0.6092 & 0.5772 & 0.7110 & 0.3057 & 0.0105 & 0.8266 & ~ \\ 
\textbf{CAVP} & 0.0000 & 0.8454 & 0.5669 & 0.6258 & 0.5230 & 0.7024 & 0.8076 & 0.7269 & 0.7465 & 0.6553 & 0.7321 & 0.2821 & 0.8127 & 0.9053 & ~ \\ 
\specialrule{1.5pt}{0pt}{0pt}
Class & gun & guzheng & hair-dryer & handpan & harmonica & harp & helicopter & hen & horse & keyboard & leopard & lion & man & marimba & ~ \\ 
\specialrule{1.5pt}{0pt}{0pt}
AVSegformer & 0.4728 & 0.7464 & 0.5995 & 0.9431 & 0.0000 & 0.7688 & 0.8334 & 0.3946 & 0.2563 & 0.8703 & 0.7653 & 0.8616 & 0.5782 & 0.9098 & ~ \\ 
TPAVI & 0.5106 & 0.7231 & 0.7612 & 0.9119 & 0.0000 & 0.7621 & 0.8379 & 0.5141 & 0.4684 & 0.8806 & 0.7856 & 0.8201 & 0.6245 & 0.9119 & ~ \\ 
\textbf{CAVP} & 0.5433 & 0.7352 & 0.6854 & 0.9427 & 0.0000 & 0.7616 & 0.8725 & 0.8698 & 0.3430 & 0.8783 & 0.7884 & 0.8958 & 0.6217 & 0.9223 & ~ \\ 
\specialrule{1.5pt}{0pt}{0pt}
Class & missile-rocket & motorcycle & mower & parrot & piano & pig & pipa & saw & saxophone & sheep & sitar & sorna & squirrel & tabla & ~ \\ 
\specialrule{1.5pt}{0pt}{0pt}
AVSegformer & 0.0000 & 0.1409 & 0.8527 & 0.3047 & 0.7714 & 0.3990 & 0.7633 & 0.4944 & 0.5863 & 0.2197 & 0.7148 & 0.4113 & 0.6475 & 0.7012 & ~ \\ 
TPAVI & 0.2028 & 0.0316 & 0.8035 & 0.2810 & 0.7535 & 0.4973 & 0.8320 & 0.6863 & 0.7137 & 0.1161 & 0.7707 & 0.8317 & 0.6656 & 0.7887 & ~ \\ 
\textbf{CAVP} & 0.0000 & 0.2981 & 0.8605 & 0.2403 & 0.8134 & 0.7346 & 0.8311 & 0.8037 & 0.7085 & 0.2074 & 0.8023 & 0.5639 & 0.7770 & 0.7387 & ~ \\ 
\specialrule{1.5pt}{0pt}{0pt}
Class & tank & tiger & tractor & train & trombone & truck & trumpet & tuba & ukulele & utv & vacuum-cleaner & violin & wolf & woman & ~ \\ 
\specialrule{1.5pt}{0pt}{0pt}
AVSegformer & 0.6493 & 0.6707 & 0.6241 & 0.9607 & 0.6450 & 0.5623 & 0.6298 & 0.9178 & 0.7482 & 0.0000 & 0.2465 & 0.5906 & 0.8749 & 0.5682 & ~ \\ 
TPAVI & 0.7202 & 0.7762 & 0.5412 & 0.9239 & 0.6158 & 0.4848 & 0.4430 & 0.8686 & 0.7479 & 0.0000 & 0.2122 & 0.7115 & 0.8528 & 0.6369 & ~ \\ 
\textbf{CAVP} & 0.8315 & 0.7188 & 0.5650 & 0.9710 & 0.7491 & 0.4830 & 0.6486 & 0.8840 & 0.8353 & 0.0000 & 0.6977 & 0.7634 & 0.9193 & 0.6123 & ~ \\ 
\specialrule{1.5pt}{0pt}{0pt}

\end{tabular}
}
\vspace{-10pt}
    
\end{table*}
\begin{table*}[!ht]
    \centering
    \caption{Class-level (FDR) audio-visual segmentation results (in \%) on AVSBench-Semantic dataset~\cite{zhou2023audio} (original resolution) with ResNet50~\cite{he2016deep} backbone. 
    }
    \vspace{-5pt}
    \label{tab:avss_class_fdr}
    \def\arraystretch{1.1}
    \resizebox{\linewidth}{!}{
    \begin{tabular}{!{\vrule width 1.5pt}c!{\vrule width 1.5pt}ccccccccccccccc!{\vrule width 1.5pt}} 
\specialrule{1.5pt}{0pt}{0pt}
Class & background & accordion & airplane & axe & baby & bassoon & bell & bird & boat & boy & bus & car & cat & cello & clarinet \\ 
\specialrule{1.5pt}{0pt}{0pt}
AVSegformer & 0.0550 & 0.0479 & 0.0502 & 0.0000 & 0.4288 & 0.3772 & 0.7397 & 0.7448 & 0.0926 & 0.9487 & 0.5460 & 0.6959 & 0.4878 & 0.3081 & 0.9300 \\ 
TPAVI & 0.0631 & 0.2274 & 0.0449 & 0.0000 & 0.3432 & 0.3230 & 0.4422 & 0.4977 & 0.1435 & 0.8376 & 0.1323 & 0.4156 & 0.2250 & 0.3140 & 0.6382 \\ 
\textbf{CAVP} & 0.0505 & 0.0437 & 0.0601 & 0.0000 & 0.3147 & 0.3791 & 0.4865 & 0.6186 & 0.1240 & 0.5693 & 0.6331 & 0.5954 & 0.4191 & 0.2809 & 0.9047 \\ 
\specialrule{1.5pt}{0pt}{0pt}
Class & clipper & clock & dog & donkey & drum & duck & elephant & emergency-car & erhu & flute & frying-food & girl & goose & guitar & ~ \\ 
\specialrule{1.5pt}{0pt}{0pt}
AVSegformer & 1.0000 & 0.2708 & 0.6018 & 0.3006 & 0.7584 & 0.2165 & 0.1045 & 0.3412 & 0.2448 & 0.5871 & 0.1603 & 0.7783 & 0.5090 & 0.1339 & ~ \\ 
TPAVI & 1.0000 & 0.2492 & 0.5280 & 0.3382 & 0.2250 & 0.3010 & 0.0993 & 0.2199 & 0.3155 & 0.2966 & 0.1332 & 0.6854 & 0.9897 & 0.1824 & ~ \\ 
\textbf{CAVP} & 1.0000 & 0.1405 & 0.3474 & 0.3350 & 0.4075 & 0.2775 & 0.2348 & 0.3099 & 0.2496 & 0.1147 & 0.1721 & 0.7238 & 0.1870 & 0.0896 & ~ \\ 
\specialrule{1.5pt}{0pt}{0pt}
Class & gun & guzheng & hair-dryer & handpan & harmonica & harp & helicopter & hen & horse & keyboard & leopard & lion & man & marimba & ~ \\ 
\specialrule{1.5pt}{0pt}{0pt}
AVSegformer & 0.4300 & 0.0935 & 0.4443 & 0.0412 & 0.0000 & 0.2514 & 0.1549 & 0.2107 & 0.7680 & 0.1360 & 0.2649 & 0.1580 & 0.4202 & 0.0822 & ~ \\ 
TPAVI & 0.2801 & 0.2203 & 0.2559 & 0.0553 & 1.0000 & 0.1932 & 0.1607 & 0.3488 & 0.5360 & 0.1038 & 0.1826 & 0.2010 & 0.3472 & 0.0798 & ~ \\ 
\textbf{CAVP} & 0.3705 & 0.2483 & 0.3653 & 0.0559 & 0.0000 & 0.2581 & 0.1288 & 0.0918 & 0.6947 & 0.1201 & 0.2287 & 0.1150 & 0.3599 & 0.0692 & ~ \\ 
\specialrule{1.5pt}{0pt}{0pt}
Class & missile-rocket & motorcycle & mower & parrot & piano & pig & pipa & saw & saxophone & sheep & sitar & sorna & squirrel & tabla & ~ \\ 
\specialrule{1.5pt}{0pt}{0pt}
AVSegformer & 1.0000 & 0.7616 & 0.0958 & 0.6423 & 0.2116 & 0.6355 & 0.1542 & 0.4672 & 0.3244 & 0.6554 & 0.1578 & 0.5089 & 0.2943 & 0.3404 & ~ \\ 
TPAVI & 0.7909 & 0.9448 & 0.1636 & 0.6326 & 0.1957 & 0.3731 & 0.1337 & 0.1972 & 0.1895 & 0.8594 & 0.1386 & 0.1214 & 0.2973 & 0.2370 & ~ \\ 
\textbf{CAVP} & 1.0000 & 0.4343 & 0.0645 & 0.7223 & 0.1387 & 0.2925 & 0.1293 & 0.1074 & 0.2643 & 0.7786 & 0.1885 & 0.4891 & 0.2233 & 0.2929 & ~ \\
\specialrule{1.5pt}{0pt}{0pt}
Class & tank & tiger & tractor & train & trombone & truck & trumpet & tuba & ukulele & utv & vacuum-cleaner & violin & wolf & woman & ~ \\ 
\specialrule{1.5pt}{0pt}{0pt}
AVSegformer & 0.1411 & 0.3576 & 0.4080 & 0.0274 & 0.2927 & 0.1022 & 0.2312 & 0.0539 & 0.2487 & 1.0000 & 0.7833 & 0.3927 & 0.1284 & 0.4196 & ~ \\ 
TPAVI & 0.1946 & 0.1879 & 0.5165 & 0.0732 & 0.3823 & 0.0581 & 0.5315 & 0.1349 & 0.2057 & 1.0000 & 0.8205 & 0.2718 & 0.1152 & 0.3276 & ~ \\ 
\textbf{CAVP} & 0.1282 & 0.3183 & 0.4969 & 0.0285 & 0.2766 & 0.1775 & 0.2415 & 0.1326 & 0.1567 & 1.0000 & 0.3237 & 0.2250 & 0.0810 & 0.3646 & ~ \\ 
\specialrule{1.5pt}{0pt}{0pt}
\end{tabular}
}
\vspace{-10pt}
    
\end{table*}

{
    \small
    \bibliographystyle{ieeenat_fullname}
    \bibliography{main}
}


\end{document}